%% file: main.tex
\documentclass[times,twocolumn,final]{elsarticle}
\usepackage{medima}
\usepackage{framed,multirow}
\usepackage{relsize}
\usepackage{amsmath}
\usepackage{mathrsfs}
\usepackage{tabularx,ragged2e}
\usepackage{makecell}
\usepackage{cite}
\usepackage{comment} 
\usepackage{booktabs}
\usepackage{amssymb}
\usepackage{latexsym}
\usepackage{pgfplots}
\usepackage{tikz}
\usepackage{threeparttable}
\usepackage{float}
\usepackage{url}
\usepackage{xcolor}
\usepackage{dblfloatfix}
\usepackage{hhline}
\usepackage{array}
\usepackage{diagbox}
\usepackage[section]{placeins}
\usepackage{booktabs}
\usepackage{multicol}
\usepackage{mathtools}
\usepackage{textcomp,xspace}
\usepackage{subfig}
\usepackage{amsfonts,bm}
\usepackage{graphicx,epstopdf}
\usepackage[normalem]{ulem}
\usepackage{tabulary}
\usepackage[colorlinks=true,citecolor=blue,urlcolor=blue]{hyperref}
\usepackage[]{color}
\usepackage{fancyhdr}
\usepackage{soul}
\usepackage{bbm}
\usepackage{pifont}

\usepackage[leftcaption]{sidecap}

\newcommand{\red} {\color{black}}

\newcolumntype{"}{@{\hskip\tabcolsep\vrule width 1pt\hskip\tabcolsep}}
\newcolumntype{P}[1]{>{\centering\arraybackslash}p{#1}}
\newcommand{\B}[1]{\textbf{#1}}
\newcommand{\Zstroke}{\text{\ooalign{\hidewidth\raisebox{0.2ex}{--}\hidewidth\cr$Z$\cr}}}
\newcommand{\triplet}[1]{\textlangle{\textit{#1}}\textrangle{}}
\newcommand{\videolink}{\url{https://youtu.be/d_yHdJtCa98}}
\newcommand*{\Scale}[2][4]{\scalebox{#1}{$#2$}}%

\definecolor{newcolor}{rgb}{.8,.349,.1}
\hypersetup{pdfborder={0 0 0},
	colorlinks=true,
	linkcolor=black,
	citecolor=black,
	urlcolor=black}
\begin{document}

\verso{Chinedu Innocent Nwoye \textit{et~al.}}

\begin{frontmatter}

\title{Rendezvous: Attention Mechanisms for the Recognition of Surgical Action Triplets in Endoscopic Videos}

\author[1]{Chinedu Innocent \snm{Nwoye}\corref{cor1}}
\ead{nwoye@unistra.fr}
\cortext[cor1]{Corresponding author: 
  Tel.: +33 (0) 3 904 13 535;}
\author[1]{Tong \snm{Yu}}
\author[2,3]{Cristians \snm{Gonzalez}}
\author[2,3]{Barbara \snm{Seeliger}}
\author[1,4]{Pietro \snm{Mascagni}}
\author[2,3]{\\Didier \snm{Mutter}}
\author[5]{Jacques \snm{Marescaux}}
\author[1,2]{Nicolas \snm{Padoy}}
\ead{npadoy@unistra.fr}

\affliation[1]{ICube, University of Strasbourg, CNRS, France\hspace{5 mm}}
\affliation[2]{IHU Strasbourg, France\hspace{5 mm}}
\affliation[3]{University Hospital of Strasbourg, France\\}
\affliation[4]{Fondazione Policlinico Universitario Agostino Gemelli IRCCS, Rome, Italy\hspace{5 mm}}
\affliation[5]{IRCAD France}

\input{00-abstract.tex}

\end{frontmatter}

\input{01-introduction}

\input{02-literature}
\input{03-dataset}
\input{04-methodology}
\input{05-experiment}

\input{06-results}

\input{07-conclusion}
\input{08-acknowledgement}
\bibliographystyle{model2-names.bst}
\biboptions{authoryear}
\bibliography{main.bbl}

\input{10-appendix}
\end{document}

%% file: 00-abstract.tex
\begin{abstract}
\begin{center}
    \rule{2.07\linewidth}{1.0pt}\\[0.15in]
\end{center}
\noindent{\bf Abstract: }
Out of all existing frameworks for surgical workflow analysis in endoscopic videos, action triplet recognition stands out as the only one aiming to provide truly fine-grained and comprehensive information on surgical activities. This information, presented as \triplet{instrument, verb, target} combinations, is highly challenging to be accurately identified. Triplet components can be difficult to recognize individually; in this task,
it requires not only performing recognition simultaneously for all three triplet components, but also correctly establishing the data association between them.
To achieve this task, we introduce our new model, the {\it Rendezvous} (RDV), which recognizes triplets directly from surgical videos by leveraging attention at two different levels.
We first introduce a new form of spatial attention to capture individual action triplet components in a scene; called {\it Class Activation Guided Attention Mechanism} (CAGAM). This technique focuses on the recognition of verbs and targets using activations resulting from instruments.
To solve the association problem, our RDV model adds a new form of semantic attention inspired by Transformer networks; {called {\it Multi-Head of Mixed Attention} (MHMA). This technique uses several} cross and self attentions to effectively capture relationships between instruments, verbs, and targets.
We also introduce {\it CholecT50} - a dataset of 50 endoscopic videos in which \textit{every} frame has been annotated with labels from 100 triplet classes.
Our proposed RDV model significantly improves the triplet prediction mean AP by over 9\% compared to the state-of-the-art methods on this dataset.

 \vspace{-0.15in}
\end{abstract}

%% file: 01-introduction.tex
\section{Introduction}\label{sec:introduction}

\begin{figure}[!ht]
\centering
\includegraphics[width=0.95\linewidth]{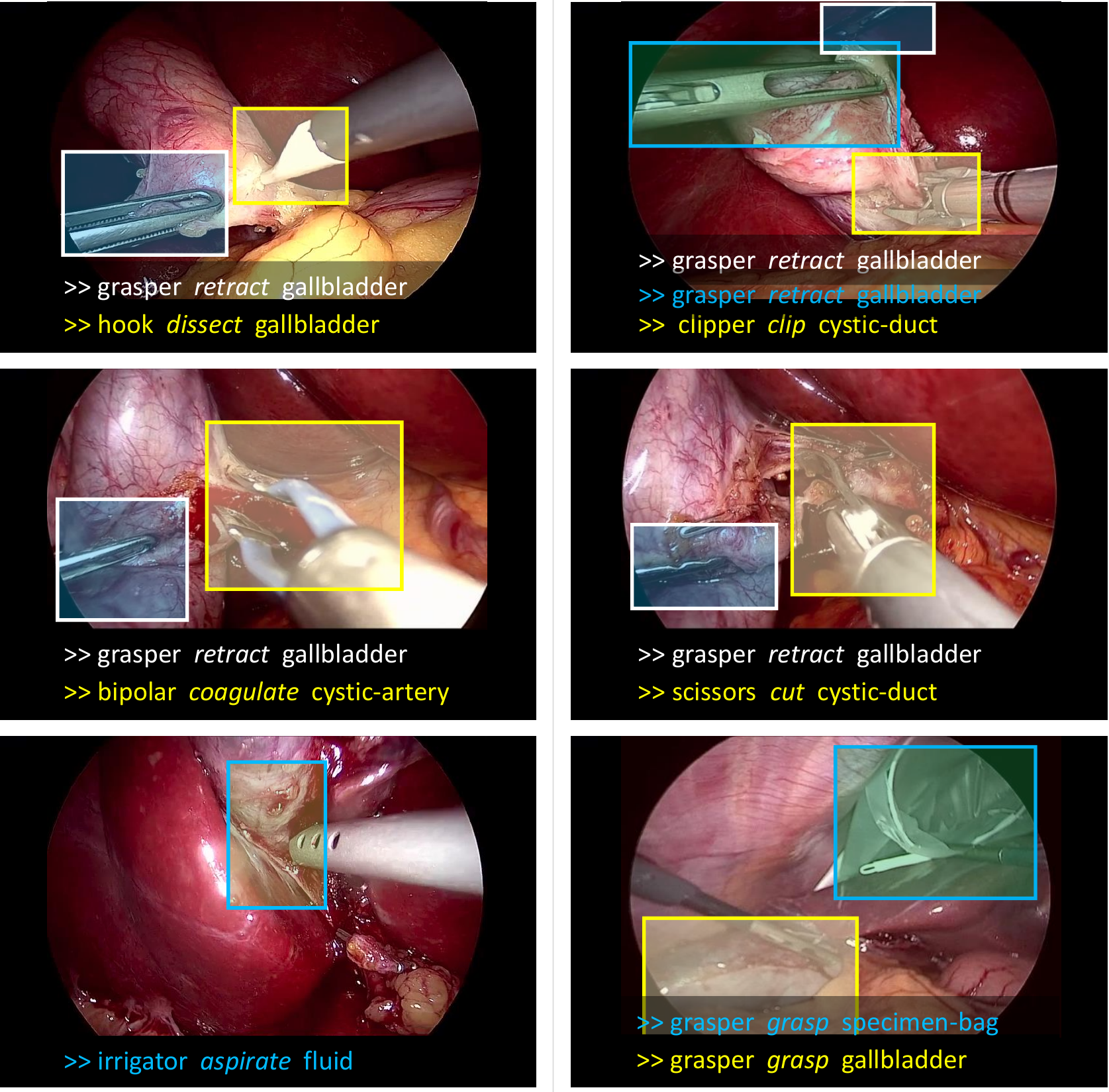} 
\caption{Some examples of action triplets from the CholecT50 dataset. The localization is not part of the dataset, but a representation of the weakly-supervised output of our recognition model.}
\label{fig:data}
\end{figure}

Laparoscopic cholecystectomy, as one of the most commonly performed surgical procedures in the world {\citep{shaffer2006epidemiology,majumder2020laparoscopic}}, has become the gold standard approach \citep{pucher2018outcome} over its open surgery counterpart.
As a minimally invasive procedure, it significantly alleviates some of the preoperative, intraoperative, and postoperative burden: the patient generally experiences decreased odds of nosocomial infection, less pain, less bleeding, and faster recovery times \citep{velanovich2000laparoscopic}. 
Yet, this success comes at a price for the surgeon, who now has to deal with increased technical difficulty coming from the indirect vision and laparoscopic instruments \citep{ballantyne2002pitfalls}, especially during complex cases \citep{felli2019feasibility}.
The elevated complexity of laparoscopy is one of the motivations driving the development of context-aware support systems for surgery \citep{maier2017surgical}; i.e. systems capable of assisting surgeons, for example via automated warnings \citep{vercauteren2019cai4cai}, 
based on their dynamic perception and understanding of the surgical scene and workflow.

Developing this understanding is the focus of surgical workflow analysis methods: \textit{given a scene from surgery, what is happening in it?} The finer-grained the answer becomes, the more value it gains in terms of clinical utility: for instance according to \citet{mascagni2021artificial}, an automated surgical safety system would benefit from the ability to identify individual actions such as a {\it clipper} applying a {\it clip} to the {\it cystic-artery} or other blood vessels.

Methods from the literature have so far only given incomplete answers, with our previous work as the only exception \citep{nwoye2020recognition}. The main task studied by the community, surgical phase recognition \citep{ahmadi2006recovery,phase_lo2003episode}, only describes scenes at a very coarse level. As an example the {\it clipping and cutting} phase \citep{twinanda_endonet_ieee2017} in cholecystectomy contains a multitude of important actions: {\it graspers} holding anatomical landmarks, a {\it clipper} applying several clips, laparoscopic {\it scissors} cutting the {\it cystic-duct} and so on. 
The phase information on its own does not, by any means, provide an accurate picture of the activities taking place.  
Even finer-grained workflow divisions such as steps \citep{ramesh2021multi} are composed of multiple individual actions.
Limited attempts were made in other works \citep{khatibi2020proposing,rupprecht2016sensor} as well as in MICCAI 2019's Endoscopic Vision challenge\footnote{\scriptsize{\url{https://endovissub-workflowandskill.grand-challenge.org/}}} to capture those actions focusing on key verbs such as {\it dissect, cut, or coagulate}. This type of framework, however, overlooks interactions with the anatomy.

To obtain a comprehensive account of a surgical scene, simultaneous recognition of the instruments, verbs, target anatomy and the relationships between them needs to be achieved. 
This goes beyond the conventional action recognition used in the EndoVis 2019 sub-challenge to a deeper understanding of visual semantics that depicts the complex relationships between instruments and tissues.
\citet{katic2014knowledge} proposed the {\it surgical action triplet} as the most detailed and expressive formalism for surgical workflow analysis. However, while \citet{katic2014knowledge,katic_lapontospm_ijcars2015} leveraged triplet formulation provided by manual annotation to better recognize surgical phases, no attempt outside of our previous work \citep{nwoye2020recognition} has been made to directly recognize those triplets {from surgical images or videos}.

Nonetheless, the difficulty of this recognition task, for all its utility, is not to be overlooked. First, action triplets are {\bf instrument-centric}: meaning that visibility alone does not determine the consideration of an anatomy as a part of a triplet, but also their involvement in an interaction carried out by an instrument. 
For instance, the {\it liver}, which is visible most of the time in laparoscopic cholecystectomy, is labeled a target only when being acted upon by an instrument. 
Similarly, a verb is defined by the instrument's action, and so, without an instrument, there cannot be a verb.
Furthermore, the level of {\bf spatial reasoning} involved is highly challenging: given an instrument, its role (verb) with respect to a given target can imperceptibly change: \triplet{grasper, retract, gallbladder}, \triplet{grasper, grasp, gallbladder}, \triplet{grasper, dissect, gallbladder} are tough to distinguish even for experienced surgeons and require careful observation of the area surrounding the tooltip.
The applications of the surgical instruments vary according to the surgeon’s intention for use.
Multiplicity and {\bf semantic reasoning} are the other major challenges. 
Overlaps are found between different instruments used for the same action (or verb), e.g. {\it dissection} performed by {\it bipolar, grasper, hook, irrigator}, and {\it scissors } (see Table \ref{table:data_stat}). 
Similarly, when operating on an organ or structure, multiple instruments can interact with the target. Thus, a target can be simultaneously involved in multiple distinct actions, e.g. \triplet{grasper, retract, cystic-duct} and \triplet{hook, dissect, cystic-duct} happening at the same time.
Since multiple triplets can occur in one frame, associating the matching components of the triplets is akin to solving a complex tripartite data association problem between the entities. The model from our previous work, as a first attempt to tackle the triplet recognition problem, addresses those challenges in a limited manner, without explicit spatial focus or an advanced enough model of the instrument - verb - target relationships.

In this paper, we extend \citet{nwoye2020recognition}, our conference paper published in MICCAI 2020. Our extension is both in data and in methods.
On the data contribution, we introduce the \textit{CholecT50} dataset, which is a quantitative and qualitative expansion of the CholecT40 dataset \citep{nwoye2020recognition}. The dataset consists of 50 videos of cholecystectomy annotated with 161K instances from 100 triplet classes. 
Examples of such action triplets include: 
\triplet{grasper, retract, gallbladder}, 
\triplet{hook, dissect, omentum},
\triplet{clipper, clip, cystic-artery}, 
\triplet{scissors, cut, cystic-duct},
\triplet{irrigator, aspirate, fluid},
etc, as also shown in Fig \ref{fig:data}.

Method-wise, we develop a new recognition model called \textit{Rendezvous} (RDV), which is a transformer-inspired neural network for surgical action triplet recognition. RDV improves on the existing \textit{Tripnet} model proposed in \citet{nwoye2020recognition} by leveraging the attention mechanism to detect the various components of the triplets and learn their association. Our first effort at exploiting attention mechanisms in this task led to the development of a \textit{Class Activation Guided Attention Mechanism} (CAGAM) to better detect the verb and target components of the triplet, which are instrument-centric.
The CAGAM is achieved by redesigning the saliency-guided attention mechanism in \citep{ji2019saliency,yao2020saliency} to utilize a more adequate and easier to learn class activation map (CAM).
While our approach is similar in the attention guiding principle, it differs in three respects: (a) our attention network is guided by the instrument's activations which are learnable in the same network, using a global pooling layer without relying on a {\it third-party} saliency generation network, (b) our attention guide implements a combination of position and channel attention for the target and verb detection tasks respectively,
(c) we employ cross-attention from the instrument domain to the other task domains (i.e.: {\it verb and target}) as opposed to self-attention in \citet{yao2020saliency}.
Meanwhile, the CAGAM is an improvement on the class activation guide (CAG) module introduced in \citet{nwoye2020recognition} which is simply a concatenation of the model's intermediary features with the instrument's activation features.
As an ablation experiment, we show the improved performance of our previous Tripnet model, only upgraded with the CAGAM. This upgraded model is called \textit{Attention Tripnet}.

The proposed RDV uses the CAGAM unit as part of its encoder and a {\it multiple heads of mixed attention} (MHMA) decoder to learn the action triplets. The MHMA in RDV is inspired by the Transformer model \citep{vaswani2017attention} used in Natural Language Processing (NLP), particularly, its multi-head attention mechanism. Unlike in the NLP Transformer, which implements a multi-head of self-attention, we design a novel multi-head attention module that is a mixture of self- and cross-attention suitable for the action triplet recognition task.
As opposed to Transformers in NLP which use multi-head attention temporally (i.e. focusing on a sequence of words in a sentence), and the Vision Transformer \citep{dosovitskiy2020image} which uses it spatially (by forming its sequence from different patches of an image), our RDV model takes a different approach by employing it semantically, with redesigned multi-head modeling attention across the discriminating features of various components that are interacting to form action triplets.
With this method, we outperform the state-of-the-art models significantly on triplet recognition.
We plan to release our source code along with the evaluation script on our public github\footnote{\url{https://github.com/CAMMA-public/tripnet}}$^,$\footnote{\url{https://github.com/CAMMA-public/rendezvous}}$^,$\footnote{\url{https://github.com/CAMMA-public/attention-tripnet}}~upon acceptance of this paper.
The dataset will also become public on the CAMMA team's website\footnote{\url{http://camma.u-strasbg.fr/datasets}}.

In summary, the contributions of this work are as follows:
\begin{enumerate}
    \item We present a comprehensive study on surgical action triplet recognition directly from videos.
    \item We propose a Class Activation Guided Attention Mechanism ({\it CAGAM}) for detecting the target and verb components of the triplets conditioned on the instrument's appearance cue. {
    \item We propose a Multi-Head of Mixed Attention ({\it MHMA}) by modeling self- and cross-attention on semantic sequences of class-wise representations to learn the interaction between the instrument, verb, and target in a surgical scene.
    \item We develop {\it Rendezvous (RDV)}: a transformer-inspired neural network model that utilizes CAGAM and MHMA for surgical triplet recognition in laparoscopic videos.}
    \item We present a large endoscopic action triplet dataset, {\it CholecT50}, for this task.
    \item We analyze the surgical relevance of our methods and results, setting the stage for clinical translation and future research. 
\end{enumerate}

%% file: 02-literature.tex
\section{Related Work}\label{sec:literature}

\subsection{Surgical Workflow Analysis} 
\label{sec:literature:workflow}
The paradigm shift brought by Artificial Intelligence (AI) across several fields has seen the application of deep learning techniques for the recognition of surgical workflow activities to provide assisted interventions in the operating room (OR). However, compared to other fields such as natural Vision, NLP, Commerce, etc., there has been a delay in introducing large-scale data science to interventional medicine.
This is partly due to the unavailability of large annotated dataset \citep{maier2017surgical} and the particular need for precision in medicine.
Some research focuses on detecting elements such as instruments/tools used during surgery \citep{mai:al2018monitoring,garcia2017toolnet,nwoye_convlstm_ijcars2019,sznitman2014fast,miccai:vardazaryan2018weakly,voros2007automatic}, while others model the sequential workflow by recognizing surgical phases either from endoscopic videos \citep{phase_blum2010modeling,phase_dergachyova2016automatic,phase_funke2018temporal,phase_lo2003episode,twinanda_endonet_ieee2017,yu2018learning,phase_zisimopoulos2018deepphase}
or from ceiling-mounted cameras \citep{chakraborty2013video,twinanda2015data}.
Some works go deeper in the level of granularity, recognizing the steps within each surgical phase \citep{charriere2014automated,lecuyer2020assisted,ramesh2021multi}, while others learn phase transitions \citep{sahu2020surgical}.
Another work \citep{lo2003episode} investigated the four major events in minimally invasive surgery (MIS) and categorized these events into their main actions; namely, {\it idle, retraction, cauterization}, and {\it suturing}.
From the perspective of robotic surgery, similar research focused more on gesture recognition from kinematic data \citep{dipietro_recognizing_miccai2016,dipietro_segmenting_ijcars2019}, and robotized surgeries \citep{kitaguchi_real_sr2019,zia_surgical_miccai2018}, system events \citep{malpani2016system}, and the recognition of other events, such as the presence of smoke or bleeding \citep{loukas2015smoke}.
These surgical events are explored for the recognition of surgeon's deviation from standard processes in laparoscopic videos \citep{huaulme2020offline}.

Aside the coarse-grained activities, some works \citep{khatibi2020proposing,rupprecht2016sensor} explored fine-grained action in laparoscopic videos, however, the recognition task is limited to verb classification.
Within the EndoVis challenge at MICCAI 2019, 
\citet{wagner2021comparative} introduced a similar action recognition task for only four prevalent verbs in surgery ({\it cut, grasp, hold,} and {\it clip}), however, this does not consider the target anatomy or the instrument performing the action.
In the SARAS-ESAD challenge organized within MIDL 2020, the proposed action labels encompass 21 classes \citep{bawa2021saras}. The EASD challenge dataset is an effort to capture more details in surgical action recognition. 
While this dataset provides spatial labels for action detection, just like some human-object interaction (HOI) datasets, it formalizes action labels as verb-anatomy relationship such as {\it clippingTissue, pullingTissue, cuttingTissue}, etc., and thus, do not take into account the instrument performing the actions.
Although humans are not categorized in the general vision HOI problem, it is imperative to recognize the surgical instruments by their categories as they play semantically different roles; their categories are informative in distinguishing the surgical phases.
Recognizing surgical actions as single verbs is also being explored in other closely related procedures such as gynecologic laparoscopy \citep{khatibi2020proposing,kletz2017surgical,petscharnig2018early}.

For a more detailed workflow analysis, \citet{nwoye2020recognition} proposed to recognize surgical actions at a fine-grained level directly from laparoscopic cholecystectomy videos, modeling them as triplets of the used instrument, its role (verb), and its underlying target anatomy. 
Such fine-grained activity recognition gives a detailed understanding of the image contents in laparoscopic videos.

\subsection{Surgical Action Triplet Recognition} 
\label{sec:literature:soa}
In the existing surgical ontology, an action is described as a triplet of the used instrument, a verb representing the action performed, and the anatomy acted upon \citep{neumuth_acquisition_icdesa2006,katic2014knowledge}. 
Earlier works such as \citet{katic2014knowledge,katic_lapontospm_ijcars2015} used triplet annotation information to improve surgical phase recognition.
Recently, \citet{nwoye2020recognition} introduced CholecT40, an endoscopic video dataset annotated with action triplets. 
Tripnet \citep{nwoye2020recognition} is the first deep learning model designed to recognize action triplets directly from surgical videos. The model relies on a class activation guide (CAG) module to detect the verb and target in triplets, leveraging instrument appearance cues. It models the final triplet association by projecting the detected components to a 3D interaction space (3Dis) to learn their association while maintaining a triplet structure.
In this paper, we improve on the verb and target detections using an attention mechanism. The triplet dataset \citep{nwoye2020recognition} is also expanded and refined.

With fine-grained action recognition now gaining momentum, a recent work in robotic surgery \citep{xu2021learning} extended two robotic surgery datasets, MICCAI's robotic scene segmentation challenge \citep{allan20202018} and Transoral Robotic Surgery (TORS), with 11 and 5 semantic relationship labels respectively.
They in turn proposed a cross-domain method for the two datasets generating surgical captions that are comparable to action triplets.

Detecting multi-object interaction in natural images/videos is widely explored by the research on human-object interaction (HOI) \citep{hu2013recognising,mallya2016learning} where activities are formulated as triplets of \triplet{human, verb, object} \citep{chao2015hico}.
Detecting or recognizing HOI is enabled by triplet datasets with spatial annotations (e.g. HICO-DET \citep{chao2018learning}, VCOCO \citep{lin2014microsoft}) or simply binary presence labels (e.g. HICO \citep{chao2015hico}).
CNN models with simple \citep{mallya2016learning} or multi-stream architectures \citep{chao2018learning} are widely used to model human and object detections as well as resolving spatial relationships between them.
Considering the often large number of possible combinations, \citet{shen2018scaling} proposed a zero-shot method to predict unseen verb-object pairs at test time.

\subsection{Attention Mechanism}\label{sec:literature:attention}
Since the advent of the attention mechanism \citep{bahdanau2014neural}, many deep learning models have exploited it in various forms:~from self \citep{vaswani2017attention} to cross \citep{mohla2020fusatnet}, and from spatial \citep{fu2019dual} to temporal \citep{sankaran2016temporal}. 
Methods relying on attention mechanisms \citep{wang2019deep,kolesnikov2019detecting} are proposed to focus the HOI detection networks only on crucial human and object context features.
An action-guided attention mining loss \citep{linaction} has also been used in HOI recognition tasks; however, all these attention models rely on expensive \textit{spatial} annotations.

Recently, \citet{ji2019saliency} proposed a form of attention that rely on saliency features without requiring additional supervision. While \citet{ji2019saliency} used a combination of spatial and textual attention modules to capture fine-grained image-sentence correlations, another work by \citet{yao2020saliency} utilized image saliency to guide an attention network for weakly supervised object segmentation.
In medical imaging, Attention U-Net \citep{oktay2018attention} is used to focus on target structures for pancreas segmentation.

Action triplets are instrument-centric: the instrument is the verb's subject, and a visible anatomical part is only considered a target if an instrument operates on it; therefore learning the verb and target are conditioned on the instrument's presence and position. \citet{nwoye2020recognition} addressed this with an activation guide layer named CAG, where the verb and target features are each attuned to instrument activation maps. 
Even for HOI detection, which is human-centric, human appearance cues are leveraged to predict action-specific densities over target object locations, albeit fully supervised on human bounding boxes \citep{gkioxari2018detecting}.
\citet{ulutan2020vsgnet} opined that attention modeling is superior to feature concatenation in terms of spatial reasoning.
We improve on the CAG principle with a class activation-guided attention mechanism (CAGAM) achieved by redesigning the saliency-guided attention mechanism in \citep{ji2019saliency,yao2020saliency} to utilize a more adequate and easier to learn class activation map (CAM). Our implementation combines both channel and position attention mechanisms for the verb and target detections respectively.

The Transformer model \citep{vaswani2017attention} introduced in NLP shows that attention can be expanded to capture long-range dependencies without recurrence.
The Vision Transformer \citep{dosovitskiy2020image} explored this technique for image understanding with encouraging performance.
Another Transformer with end-to-end self-attention \citep{zou2021end,kim2021hotr} modeled long-range attentions for both HOI components detection and their interaction association.
In surgical data science, transformers have been explored for surgical instrument classification \citep{kondo2020lapformer} and recently for phase recognition \citep{gao2021trans,czempiel2021opera}.
Similarly, we propose {\it Rendezvous} (RDV), a transformer-inspired method, for online surgical action triplet recognition. 
The novelty of RDV is found in the powerful way it incorporates self- and cross-attentions in its multi-head layers to decode the interactions between the detected instruments and tissues in a laparoscopic procedure.

The Transformer, as used in Natural Language Processing \citep{vaswani2017attention}, learns attention maps over a {\it temporal sequence}, considering a sentence to be a sequence of words. 
In computer vision, many works have tried to replicate this by modeling input sequence over temporal frames \citep{girdhar2019video}. A single image, however, can be as informative as a complete sentence. Even the Vision Transformer \citep{dosovitskiy2020image} shows that an image is equivalent to $16\times16$ words, modeled as a {\it sequence of patches} from a single image. Many works have followed similar approaches in image understanding \citep{dosovitskiy2020image}, object detection \citep{carion2020end}, segmentation \citep{chen2021transunet,valanarasu2021medical}, captioning \citep{liu2021cptr,sundaramoorthy2021end}, and activity recognition \citep{bertasius2021space,gavrilyuk2020actor}. Alternatively, the hybrid architecture of the Vision Transformer \citep{dosovitskiy2020image} shows that, aside from the raw image, the input sequence can also be obtained from CNN features. It also shows that a patch can have a $1\times 1$ spatial size which is akin to using an image with no explicit sequence modeling.
We propose another hybrid approach to obtain the appropriate feature, one that can preserve the spatial and class-wise relationships of the interacting triplet components in a surgical image frame, a {\it semantic sequence} in this regard. 
While our attention input features are extracted from a CNN as done in the hybrid Vision Transformer, 
our attention is design to leverage the CNN's features in a manner that helps the attention network benefit from the learned class representations. This means we preserve the spatial relationship in the grid of features without breaking it up into patches.
This provides further insight into the decision-making of attention networks. 

%% file: 03-dataset.tex
\section{CholecT50: Cholecystectomy Action Triplet Dataset} \label{sec:dataset}
CholecT50 is a dataset of endoscopic videos of laparoscopic cholecystectomy surgery introduced to enable research on fine-grained action recognition in laparoscopic surgery.
It is annotated with triplet information in the form of \triplet{instrument, verb, target}.
The dataset is a collection of 50 videos consisting of 45 videos from the Cholec80 dataset \citep{twinanda_endonet_ieee2017} and 5 videos from an in-house dataset of the same surgical procedure.
It is an extension of CholecT40 \citep{nwoye2020recognition} with 10 additional videos and standardized classes.

The cholecystectomy recordings were annotated by two surgeons using the software {\it Surgery Workflow Toolbox-Annotate} from the B-com institute\footnote{\url{https://b-com.com/}}. Annotators set the beginning and end on a timeline for each identified action, then assigned to the corresponding \textit{instrument}, \textit{verb} and \textit{target} class labels.
An action ends when the corresponding instrument exits the frame, or if the verb or target changes. Out-of-frame actions are not reported, and video frames that are recorded outside the patient's body are zeroed out.

We then define classes for the triplet. Due to the number of instruments, verbs, and targets available, the theoretical number of all possible triplet configurations (900) is prohibitively high. 
Even limiting those configurations to the approximately 300 observed in the dataset has little clinical relevance due to the presence of many spurious classes.
To have a reasonable number of classes with maximum clinical utility, a team of clinical experts selected the top relevant labels for the triplet dataset. This is achieved in two steps.
In the first instance, class grouping ($\cup$) is carried out to super-class triplets that are semantically the same. 
Some examples of triplets grouped include:

{\small
\begin{enumerate}
    \item  \triplet{irrigator, aspirate, bile} $\cup$ \triplet{irrigator, aspirate, fluid} $\cup$ \triplet{irrigator, aspirate, blood}  $\longrightarrow$ \triplet{irrigator, aspirate, fluid}
    \item \triplet{grasper, pack, gallbladder} $\cup$ \triplet{grasper, store, gallbladder} $\longrightarrow$ \triplet{grasper, pack, gallbladder}
    \item \triplet{grasper, retract, gut} $\cup$ \triplet{grasper, retract, duodenum} $\cup$ \triplet{grasper, retract, colon} $\longrightarrow$ \triplet{grasper, retract, gut} 
    \item \triplet{bipolar, coagulate, liver} $\cup$ \triplet{bipolar-grasper, coagulate, liver} $\cup$ \triplet{bipolar, coagulate, liver-bed}  $\longrightarrow$ \triplet{bipolar, coagulate, liver}
    \item \triplet{grasper, grasp, gallbladder-fundus} $\cup$ \triplet{grasper, grasp, gallbladder-neck} $\cup$ \triplet{grasper, grasp, gallbladder} $\cup$ \triplet{grasper, grasp, gallbladder-body} $\longrightarrow$ \triplet{grasper, grasp, gallbladder} 
\end{enumerate}
}

\input{tables/stats-components}

In addition to class grouping, surgical relevance rating and label mediation of the annotated data are carried out by three clinicians.
For the rating, the clinicians assigned a score from a range of [1-5] to each triplet composition based on their possibility and usefulness in the considered procedure. Their average scores, as well as the triplet's number of occurrences, is used to order the triplet classes, after which the top relevant classes are selected.
Moreso, the third clinician performed label mediation in the case of label disagreement.

\input{tables/stats-triplets}

The final dataset comprises 100 triplet classes that follow the format of \triplet{instrument, verb, target}.
The triplets are composed from 6 instruments, 10 verbs and 15 target classes, presented with their instance counts in Table \ref{table:comp_stat}. 
We present the CholecT50 dataset triplet labels including their number of occurrences in Table \ref{table:data_stat}.
We also present the co-occurrence statistics for \triplet{instrument, target} and \triplet{instrument, verb} pairs within triplets in the supplementary material.

\input{tables/stats-datasplits}

For our experiment, we down-sampled the videos to $1$fps yielding 100.86K frames annotated with $161$K triplet instances.
The video dataset is split into training, validation, and testing sets as in Table \ref{table:data_split}. The videos in the dataset splits are distributed in the same ratio to include annotations from each surgeon.

%% file: tables/stats-components.tex
\begin{table}[!t]
\centering
\caption{Statistics of the triplet's component labels in the dataset}
\label{table:comp_stat}
\resizebox{\columnwidth}{!}{%
\begin{tabular}{@{}lrclrclr@{}}
\toprule
\multicolumn{2}{c}{instrument} & \phantom{abc} &
\multicolumn{2}{c}{Verb} & \phantom{abc} &
\multicolumn{2}{c}{Target} \\ \cmidrule{4-5} \cmidrule{7-8} \cmidrule{1-2}
Label & Count &  & Label & Count &  & Label & Count \\ \midrule
bipolar & 6697 &  & aspirate & 3122 &  & abd-wall/cavity & 847 \\
clipper & 3379 &  & clip & 3070 &  & adhesion & 228 \\
grasper & 90969 &  & coagulate & 5202 &  & blood-vessel & 416 \\
hook & 52820 &  & cut & 1897 &  & cystic-artery & 5035 \\
irrigator & 5005 &  & dissect & 49247 &  & cystic-duct & 11883 \\
scissors & 2135 &  & grasp & 15931 &  & cystic-pedicle & 299 \\
 &  &  & irrigate & 572 &  & cystic-plate & 4920 \\
 &  &  & null-verb & 10841 &  & fluid & 3122 \\
 &  &  & pack & 328 &  & gallbladder & 87808 \\
 &  &  & retract & 70795 &  & gut & 719 \\
 &  &  &  &  &  & liver & 17521 \\
 &  &  &  &  &  & null-target & 10841 \\ 
 &  &  &  &  &  & omentum & 9220 \\
 &  &  &  &  &  & peritoneum & 1227 \\
 &  &  &  &  &  & specimen-bag & 6919 \\ \bottomrule
\end{tabular}}
\end{table}

%% file: tables/stats-triplets.tex
\begin{table*}[!t]
\centering
\caption{Dataset statistics showing the number of occurrences of the triplets}
\label{table:data_stat}
\resizebox{1.0\textwidth}{!}{%
\begin{tabular}{@{}lrclrclr@{}}\toprule
Name & Count && Name & Count && Name & Count \\ \cmidrule{1-2} \cmidrule{4-5} \cmidrule{7-8}
bipolar,coagulate,abdominal-wall/cavity & 434 &  & grasper,grasp,cystic-artery & 76 &  & hook,dissect,gallbladder & 29292 \\ 
bipolar,coagulate,blood-vessel & 251 &  & grasper,grasp,cystic-duct & 560 &  & hook,dissect,omentum & 3649 \\ 
bipolar,coagulate,cystic-artery & 68 &  & grasper,grasp,cystic-pedicle & 26 &  & hook,dissect,peritoneum & 337 \\ 
bipolar,coagulate,cystic-duct & 56 &  & grasper,grasp,cystic-plate & 163 &  & hook,null-verb,null-target & 4397 \\ 
bipolar,coagulate,cystic-pedicle & 77 &  & grasper,grasp,gallbladder & 7381 &  & hook,retract,gallbladder & 479 \\ 
bipolar,coagulate,cystic-plate & 410 &  & grasper,grasp,gut & 33 &  & hook,retract,liver & 179 \\ 
bipolar,coagulate,gallbladder & 343 &  & grasper,grasp,liver & 83 &  & irrigator,aspirate,fluid & 3122 \\ 
bipolar,coagulate,liver & 2595 &  & grasper,grasp,omentum & 207 &  & irrigator,dissect,cystic-duct & 41 \\ 
bipolar,coagulate,omentum & 262 &  & grasper,grasp,peritoneum & 380 &  & irrigator,dissect,cystic-pedicle & 89 \\ 
bipolar,coagulate,peritoneum & 73 &  & grasper,grasp,specimen-bag & 6834 &  & irrigator,dissect,cystic-plate & 10 \\ 
bipolar,dissect,adhesion & 73 &  & grasper,null-verb,null-target & 4759 &  & irrigator,dissect,gallbladder & 29 \\ 
bipolar,dissect,cystic-artery & 187 &  & grasper,pack,gallbladder & 328 &  & irrigator,dissect,omentum & 100 \\ 
bipolar,dissect,cystic-duct & 183 &  & grasper,retract,cystic-duct & 469 &  & irrigator,irrigate,abdominal-wall/cavity & 413 \\ 
bipolar,dissect,cystic-plate & 54 &  & grasper,retract,cystic-pedicle & 41 &  & irrigator,irrigate,cystic-pedicle & 29 \\ 
bipolar,dissect,gallbladder & 353 &  & grasper,retract,cystic-plate & 1205 &  & irrigator,irrigate,liver & 130 \\ 
bipolar,dissect,omentum & 176 &  & grasper,retract,gallbladder & 48628 &  & irrigator,null-verb,null-target & 573 \\ 
bipolar,grasp,cystic-plate & 8 &  & grasper,retract,gut & 686 &  & irrigator,retract,gallbladder & 30 \\ 
bipolar,grasp,liver & 95 &  & grasper,retract,liver & 13646 &  & irrigator,retract,liver & 350 \\ 
bipolar,grasp,specimen-bag & 85 &  & grasper,retract,omentum & 4422 &  & irrigator,retract,omentum & 89 \\ 
bipolar,null-verb,null-target & 632 &  & grasper,retract,peritoneum & 289 &  & scissors,coagulate,omentum & 17 \\ 
bipolar,retract,cystic-duct & 8 &  & hook,coagulate,blood-vessel & 57 &  & scissors,cut,adhesion & 155 \\ 
bipolar,retract,cystic-pedicle & 9 &  & hook,coagulate,cystic-artery & 10 &  & scissors,cut,blood-vessel & 21 \\ 
bipolar,retract,gallbladder & 32 &  & hook,coagulate,cystic-duct & 41 &  & scissors,cut,cystic-artery & 613 \\ 
bipolar,retract,liver & 164 &  & hook,coagulate,cystic-pedicle & 15 &  & scissors,cut,cystic-duct & 808 \\ 
bipolar,retract,omentum & 69 &  & hook,coagulate,cystic-plate & 9 &  & scissors,cut,cystic-plate & 20 \\ 
clipper,clip,blood-vessel & 51 &  & hook,coagulate,gallbladder & 217 &  & scissors,cut,liver & 90 \\ 
clipper,clip,cystic-artery & 1097 &  & hook,coagulate,liver & 189 &  & scissors,cut,omentum & 27 \\ 
clipper,clip,cystic-duct & 1856 &  & hook,coagulate,omentum & 78 &  & scissors,cut,peritoneum & 56 \\ 
clipper,clip,cystic-pedicle & 13 &  & hook,cut,blood-vessel & 15 &  & scissors,dissect,cystic-plate & 12 \\ 
clipper,clip,cystic-plate & 53 &  & hook,cut,peritoneum & 92 &  & scissors,dissect,gallbladder & 52 \\ 
clipper,null-verb,null-target & 309 &  & hook,dissect,blood-vessel & 21 &  & scissors,dissect,omentum & 93 \\ 
grasper,dissect,cystic-plate & 78 &  & hook,dissect,cystic-artery & 2984 &  & scissors,null-verb,null-target & 171 \\ \cline{7-8}
grasper,dissect,gallbladder & 644 &  & hook,dissect,cystic-duct & 7861 &  &  &  \\ 
grasper,dissect,omentum & 31 &  & hook,dissect,cystic-plate & 2898 &  & Total & 161005 \\ \bottomrule
\end{tabular}
}
\end{table*}

%% file: tables/stats-datasplits.tex
\begin{table}[h]
\centering
\centering
\caption{Statistics of the dataset split}
\label{table:data_split}
\setlength{\tabcolsep}{6pt}
\resizebox{1.0\columnwidth}{!}{%
\begin{tabular}{@{}lccrcrcr@{}}
\toprule
Data split & \phantom{abc} & \phantom{abc} & Videos & \phantom{abc} & Frames & \phantom{abc} & Labels\\ \midrule
Training &&& 35  &&  72815  &&  113884\\ 
Validation  &&& 5  &&  6797  &&  10267\\ 
Testing  &&& 10  &&  21251  &&  36854\\ \midrule
Total &&& 50  &&  100863	 &&  161005 \\\bottomrule
\end{tabular}
}
\end{table}

%% file: 04-methodology.tex
\section{Methodology}
\label{sec:methodology}

\begin{figure}[t!]
\centering
    \includegraphics[width=\linewidth]{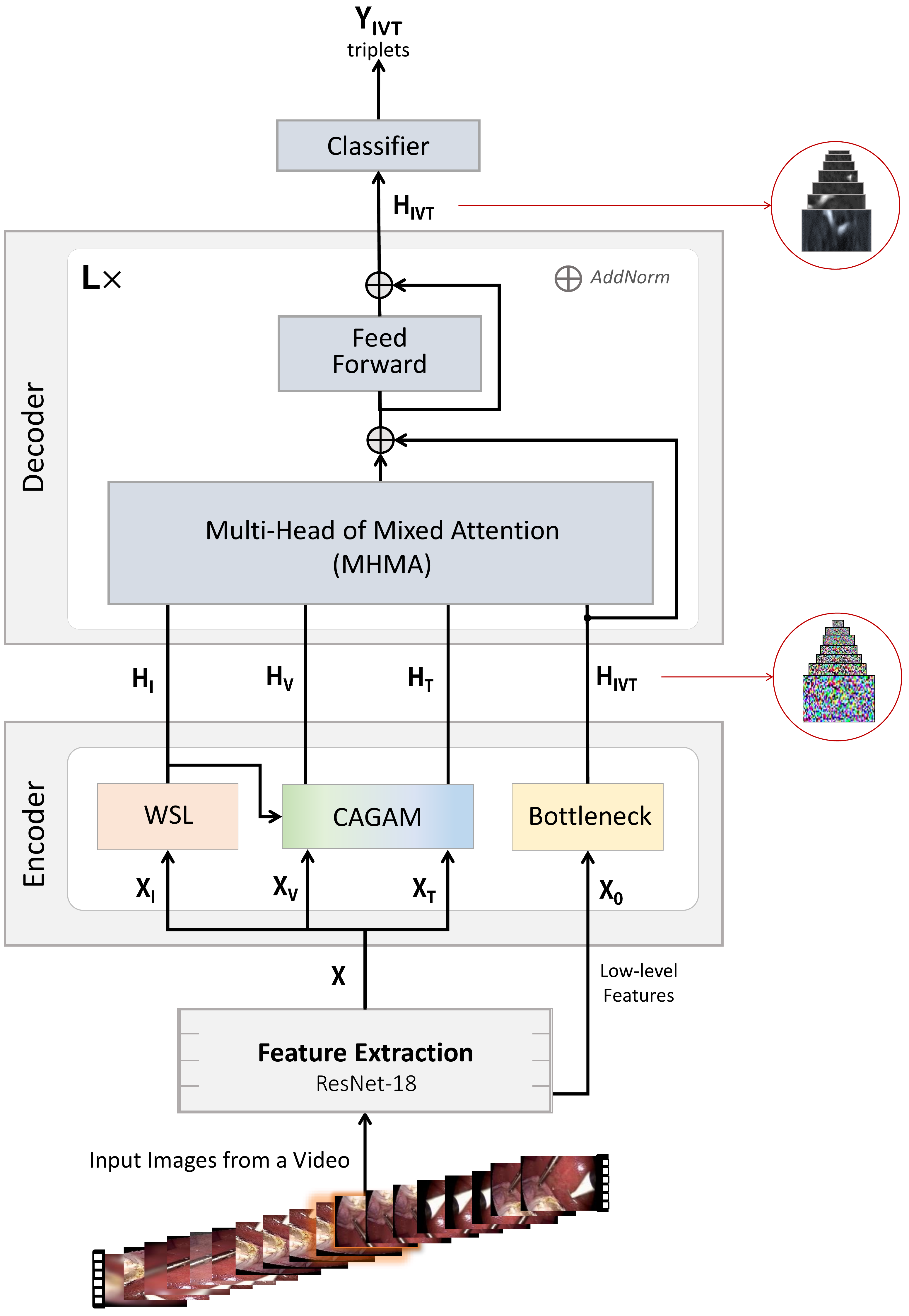}
\caption{Architecture of the Rendezvous: 
\textit{meeting of attentions}, a transformer-inspired model
with channel and position spatial attention for triplet components detection and a multi-head of self amd cross semantic attention for action triplet recognition.}
\label{fig:architecture:Rendezvous}
\end{figure}

Action triplet recognition is a complex and challenging task, since it requires: (1) simultaneously solving three multi-label classification problems, and (2) performing associations while accounting for multiple triplet instances. 
In this work, we propose two methods that tackle each aspect of these tasks.

We address the first point with the \textit{class activation guided attention mechanism} or CAGAM, which explicitly uses tool type and location information to highlight discriminative features for verbs and targets respectively. We demonstrate its utility by replacing our previous Tripnet \citep{nwoye2020recognition} model's class activation guide (CAG) with CAGAM, resulting in a preliminary model which we call {\it Attention Tripnet}.

The second point is addressed {by the {\it multi-head of mixed attention} (MHMA),} as an advanced model of semantic attention for triplet association, and a successor to the previous state-of-the-art Tripnet's more primitive 3D interaction space (3Dis). 
The MHMA resolves the triplet's components association using multiple heads of self and cross attention mechanisms.

Our final model is called the {\it Rendezvous} (RDV): a transformer-inspired neural network for surgical action triplet recognition.
The RDV combines the CAGAM in its encoder with the MHMA in its Transformer-inspired decoder for enhanced triplet component detection and association respectively.
This model provides the highest performance on action triplet recognition.

The proposed RDV network is conceptually divided into four segments: feature extraction backbone, encoder, decoder, and classifier as shown in Fig. \ref{fig:architecture:Rendezvous}.

\subsection{Feature Extraction}
\label{sec:methodology:base}
We model the visual feature extraction using the ResNet-18 base model. Our choice is motivated by the excellent performance of residual networks in visual object classification tasks. 
To facilitate more precise localization, the strides of the last two blocks of the ResNet are lowered to one pixel providing higher output resolution.

The Resnet-18 base model takes an RGB image frame from a video as input and extracts its visual features $\B{X} \in \mathbb{R}^{32\times 56\times 512}$.
The extracted feature is triplicated into ($\B{X}_I, \B{X}_V, \B{X}_T$) for multitask learning of the instrument, verb, and target components of the triplets respectively.

\subsection{Components Encoding}
\label{sec:methodology:encoder}
The encoder is responsible for detecting the various components of the triplets, while the decoder resolves the relationships between them. 
The encoder is composed of the weakly-supervised localization (WSL) module for instrument detection, class activation guided attention mechanism (CAGAM) module for verb and target recognition, and a bottleneck layer collecting unfiltered low-level features from Resnet-18's lower layer.

\subsubsection{Weakly Supervised Localization (WSL)}
\label{sec:methodology:encoder:wsl}
While this work primarily focuses on \textit{recognizing} surgical action triplets, \textit{localizing} actions -similarly to HOI tasks- is an interesting addition.
We therefore go beyond simply detecting the presence of surgical instruments by locating their position, which represents the region of interaction.
In the absence of spatial annotations we achieve this with weak supervision.

\begin{figure}[t!]
\centering
    \includegraphics[width=0.80\linewidth]{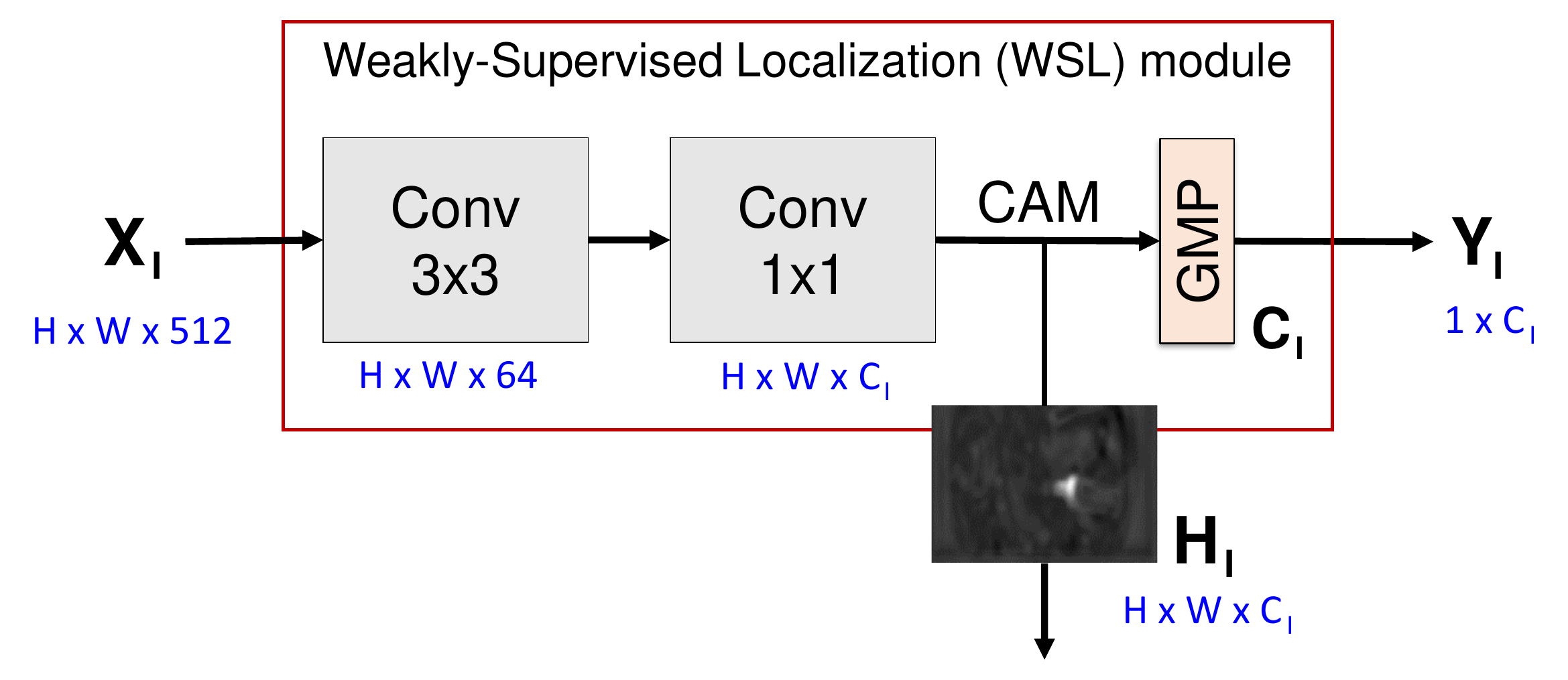}
\caption{Weakly supervised localization (WSL) layer for instrument detection. {\it Feature dimension (height $H=32$, width $W=56$, depth (class size) $C_I=6$)}.}
\label{fig:architecture:wsl}
\end{figure}

As shown in Fig. \ref{fig:architecture:wsl}, the WSL module consists of a $3\times3$ convolution layer (Conv) of 64 channels, then followed by a $1\times1$ Conv of $C_I=6$ channels for instrument localization in form of class activation maps (CAM). 

Specifically, the WSL module takes $\B{X}_I$ from the feature extraction layer as input and returns the instruments' CAM, marked as ($\B{H}_I$), from its last Conv layer.
{\red The output CAM ($\B{H}_I$) are trained for localization via their Global Maximum Pooled (GMP) values $\B{Y}_I$ representing instrument class-wise presence probabilities similar to \citet{miccai:vardazaryan2018weakly}. }

The discriminative CAM features ($\B{H}_I$) alongside these remaining extracted features ($\B{X}_V$, $\B{X}_T$) are passed to the CAGAM for verb and target detection.

\subsubsection{Class Activation Guided Attention Mechanism (CAGAM)}  
\label{sec:methodology:decoder:cagam}

Surgical action triplets are instrument-centric. Detecting the correct verbs and target anatomies is very challenging, because the visibility as well as the subtly involvement of a tool and anatomy in an action have to be taken into consideration.

A limited effort is made in our previous method, Tripnet \citep{nwoye2020recognition}, to handle this using a CAG module conditioning the detection of verbs and targets on the instruments activations, via concatenated features.
Since attention modeling is found to be superior to feature concatenation \citep{ulutan2020vsgnet}, we explore several types of attention and propose a new form of spatial attention, named CAGAM.

\begin{figure}[t!]
\includegraphics[width=1.0\linewidth]{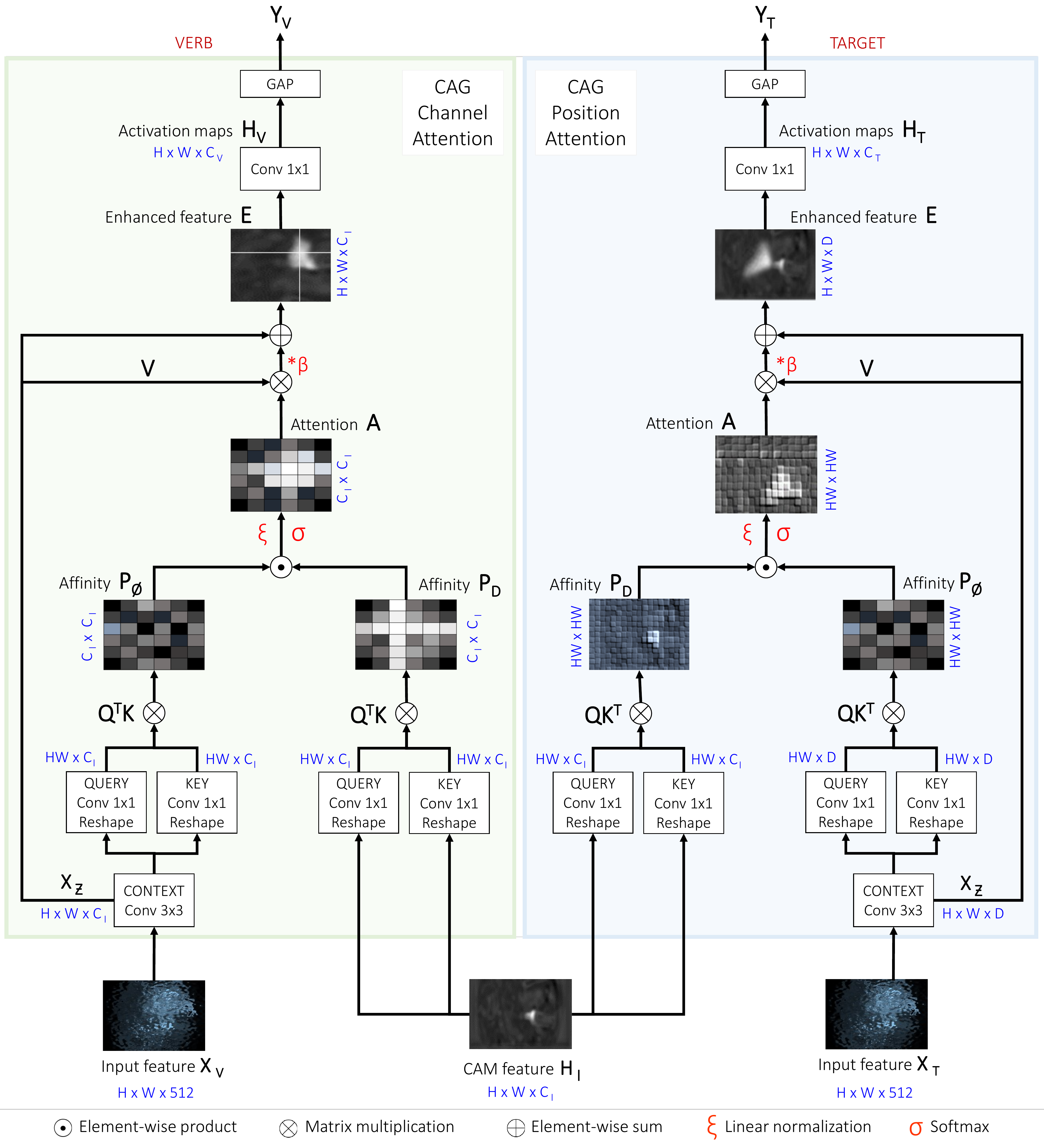}
\caption{Class Activation Guided Attention Mechanism (CAGAM):  uses the attention learned from the instrument's CAM to highlight the verb class (left) and the anatomy in contact with the instrument (right). {\it Feature dimension (height $H=32$, width $W=56$, depth $D=64$, instrument's class size $C_I=6$, verb's class size $C_V=10$, target's class size $C_T=15$).}
}
\vspace{-1mm}
\label{fig:architecture:cagam}
\end{figure}

According to \citet{vaswani2017attention}, an attention function can be described as matching a query ($\B{Q}$) and a set of key-value ($\B{K, V}$) pairs to form an output. The output is computed as a weighted sum of the values ($w\B{V}$), where the weight ($w=\B{QK}^T$) is computed by an affinity score function of the query with the corresponding key.
CAGAM is a new form of spatial attention mechanism that propagates attention from known to unknown context features, thereby enhancing the unknown context for relevant pattern discovery.
It is an adaptation of the saliency-guided attention mechanism in \citep{ji2019saliency,yao2020saliency} to utilize a more adequate and easier to learn class activation map (CAM) suitable for action triplet recognition.
It is used, in this case, to discover the verbs and targets that are involved in tool-tissue interactions leveraging the instrument's contextual dependencies, by propagating attention from the discriminative $\B{H}_I$ to the non-discriminative $\B{X}_V$ and $\B{X}_T$ features. 
The CAM ($\B{H}_I$) serves as the known context features in this regard, since they are already discriminated class-wise and localized for the instruments.

As shown in Fig. \ref{fig:architecture:cagam}, we model the CAGAM to enhance the verb's and target's unfiltered features by element-wise addition of an {\it enhancement}: this enhancement is a computed spatial attention {\bf A} from the instrument affinity maps (${\bf P}_D$) as well as the component affinity maps (${\bf P}_\emptyset$) themselves.
The ${\bf P}_D$ are termed discriminative because they originate from the instrument CAM features, whereas ${\bf P}_\emptyset$ are termed non-discriminative because they are formed from the unfiltered component features.

We observe that verbs and targets behave differently with regards to their instrument; that is, verbs are mostly affected by the instrument's type, while targets tend to be determined by instrument's position. This distinction is a key factor in the choices of attention mechanism in the CAGAM which indeed combines \textbf{channel attention} for verb detection (Fig. \ref{fig:architecture:cagam}: left) and \textbf{position attention} for target detection (Fig. \ref{fig:architecture:cagam}: right). 
Both types of spatial attention mechanisms are similar, except for the dimensions used, and therefore the nature of the information attended to. 
{The channel attention is captured in the $C_I\times C_I$ channel dimensions, informed by instrument type, whereas the position attention is captured in the ${HW\times HW}$ spatial dimensions, informed by instrument location.}
This choice is well validated in ablation studies shown further (Table \ref{tab:results:qunatitative:ablation:attention}).


\subsection*{\bf CAG channel attention for verbs}
\label{sec:methodology:attention:position}
As illustrated in Fig.~\ref{fig:architecture:cagam}~(left), verb features are first remapped to $\B{X}_\Zstroke \in\mathbb{R}^{H\times W\times C_I}$, which we call the \textit{context features}. Following two separate $1\times1$ Conv and reshapings, mapping it to a query \B{Q} and a key \B{K} of size $HW \times C_I$, the \textit{non-discriminative affinity map} $\B{P}_\emptyset\in\mathbb{R}^{C_I\times C_I}$ is obtained via matrix multiplication of the transposed \B{Q} by \B{K} as illustrated in Equation \ref{eqn:affinity}:
\begin{equation}
    {\bf P}_\emptyset = {\bf Q}^T{\bf K}.
    \label{eqn:affinity}
\end{equation}

Applying a similar process to the CAM results in the \textit{discriminative affinity map}: $\B{P}_D\in\mathbb{R}^{C_I\times C_I}$. 
As done in \citet{yao2020saliency}, an element-wise product of the two affinity maps, scaled by a factor $\xi$ and passed through {red \textit{softmax} ($\sigma$)} gives the attention $\B{A}$:
\begin{equation}
    {\bf A} = \sigma\left(\frac{{\bf P}_D{\bf P}_\emptyset}{\xi}\right).
\end{equation}

Meanwhile, we obtain the value features $\B{V}\in\mathbb{R}^{HW\times C_I}$ by reshaping the verb context $\B{X}_\Zstroke$ to $\mathbb{R}^{HW\times C_I}$.
Next, we obtain an enhancement by matrix multiplication of $\B{A}$ by $\B{V}$, weighted by a learnable temperature $\beta$.
This enhancement is reshaped to $\mathbb{R}^{H\times W\times C_I}$ and added back to $\B{X}_\Zstroke$ to produce the enhanced features, $\B{E}$.
\begin{equation}
    {\bf E} = \beta({\bf VA}) + {\bf X}_\Zstroke.
\end{equation}

The features $\B{E}$ are transformed into per-verb activation maps $\B{H}_V \in\mathbb{R}^{H\times W\times C_V}$ via a $1\times1$ Conv. Finally, verb logits $\B{Y}_{T} \in \mathbb{R}^{1\times C_V}$ are obtained by global average pooling of $\B{H}_V$, where $C_V=10$ is the number of verb classes.


\subsection*{\bf CAG position attention {\red for targets}}
\label{sec:methodology:attention:position}
{As illustrated in Fig.~\ref{fig:architecture:cagam}~(right), obtaining the $\B{Q}$, $\B{K}$, and $\B{V}$ terms for the CAG position attention is similar to the CAG channel attention mechanism.}
However to obtain an instrument location-aware attention, we multiply $\B{Q}$ by $\B{K}^T$ (instead of $\B{Q}^T$ by $\B{K}$ as done for verbs in Equation \ref{eqn:affinity}) producing affinity maps ($\B{P}_D, \B{P}_\emptyset$) and a subsequent attention map \B{A} of the desired size $HW\times HW$, informed by instrument position rather than instrument type.

Furthermore, we obtain enhanced target features ($\B{E}$), which we also feed to a $1\times1$ Conv of $C_T=15$ channels to obtain the per-target activation maps $\B{H}_T \in\mathbb{R}^{H\times W\times C_T}$. 
Using a global pooling on $\B{H}_T$, we then obtain the target logits $\B{Y}_{T} \in \mathbb{R}^{1\times C_T}$.

To ensure that the $\B{H}_I,\B{H}_V$ and $\B{H}_T$ class maps properly capture their corresponding components, we train their global pooled logits ($\B{Y}_I, \B{Y}_V, \B{Y}_T$) as auxiliary classification tasks.

\subsubsection{Bottleneck Layer}
\label{sec:methodology:encoder:bottleneck}
{In addition to these refined, component-specific features ($\B{H}_I, \B{H}_V, \B{H}_T$), a global context feature is also necessary for modeling their contextual relationship;} 
which is why we also draw a unfiltered low-level feature $\B{X}_0$ from the first block of ResNet and feed it to the bottleneck layer that consists of $3\times3\times256$ and $1\times1\times C$ convolution layers, where $C=100$ is the number of triplet classes. 
This gives the global context feature for triplets $\B{H}_{IVT}$, with channels matched to the triplet classes.

The unfiltered triplets feature $\B{H}_{IVT}$ as well as the triplet component's class maps ($\B{H}_I, \B{H}_V, \B{H}_T$) are fed to the decoder layer for decoding the triplet association.

\subsection{Interaction Decoding}
\label{sec:methodology:decoder}
{We describe here the modeling of the triplet components' relationship in the RDV decoder. 
An existing approach attempts to model every triplet possibility from the outer product combination of the three components using a 3D feature space \citep{nwoye2020recognition}.
This design models more than the required triplets, including irrelevant and impossible combinations, making the module hard to train.
Hence, we follow a Transformer-like architecture \citep{vaswani2017attention,dosovitskiy2020image,chen2021transunet} leveraging long-range attention to efficiently model the required relationships.
To take into consideration the constituting components of the triplets \citep{nwoye2020recognition}, we utilize the semantic features of each component, captured in their class maps ($H_I, H_V, H_T$).
Unlike the Vision Transformer \citep{dosovitskiy2020image}, however, we do not break class maps into patches. As shown by ablation results in the supplementary material, the patch sequence degrades representations, especially information on instruments that is important for locating actions.}

Hence, we model the RDV attention decoder on the semantic sequence of learnt class-wise representations.
From  $\B{H}_I, \B{H}_V, \B{H}_T$ and the global triplet feature $\B{H}_{IVT}$, RDV decodes all the self- and cross-interactions between the triplet's global context feature and the three features corresponding to individual components, using scaled dot-product attention \citep{vaswani2017attention} without using recurrence. In addition to self-attention, cross-attention adds the capability to better model the relationships with components participating in the action triplet. This is important when resolving interactions: for instance, an anatomical part can appear in the frame without being a target, often making the interaction with the instrument ambiguous.

To understand the attention decoder used in this work, we explain the {\bf decoding-by-attention} concept below:
\begin{enumerate}
    \item Firstly, attention decoding is described as a search process whereby a query ($\B{Q}$), that is issued by a user ({\it sink} or {\it receiver}), is used to retrieve data from a repository ({\it source}). Normally, $\B{Q}$ is a user's abridged description of the requested data also known as {\it search terms}. 
    \item The source context consists of a key-value ($\B{K,V}$) pair where $\B{V}$ is a collection of several data points or {\it records} and $\B{K}$ is the mean descriptor for each record also known as {\it keywords}. 
    \item To retrieve the requested data, the issued $\B{Q}$ is matched with the available $\B{K}$s to create an {\it affinity} ($\B{P}$), also known as the {\it attention weight}.
    \item The $\B{P}$, when matched with $\B{V}$, creates an \textit{attention map} ($\B{A}$) which helps retrieve the most appropriate data to the sink.
\end{enumerate}

We implement a transformer-inspired decoder that is composed of a stack of $L=8$ identical layers as shown in Fig. \ref{fig:architecture:Rendezvous}.
Each layer receives the triplet features $\B{H}_{IVT}$ and the encoded class maps ($\B{H}_I, \B{H}_V, \B{H}_T$) as inputs which are processed successively by its two internal modules: MHMA and feed-forward, to produce refined triplet features, $\B{H}_{IVT}$.
The output of each module is followed by a residual connection and a layer normalization (AddNorm) as it is done in other multi-head attention networks.
The entire cycle repeats, with a more refined $\B{H}_{IVT}$ output, until the $L^{th}$ layer.

\subsubsection{Multi-head of Mixed Attention (MHMA)} \label{sec:methodology:decoder:mhma}

\begin{figure}[t!]
\includegraphics[width=1.0\linewidth]{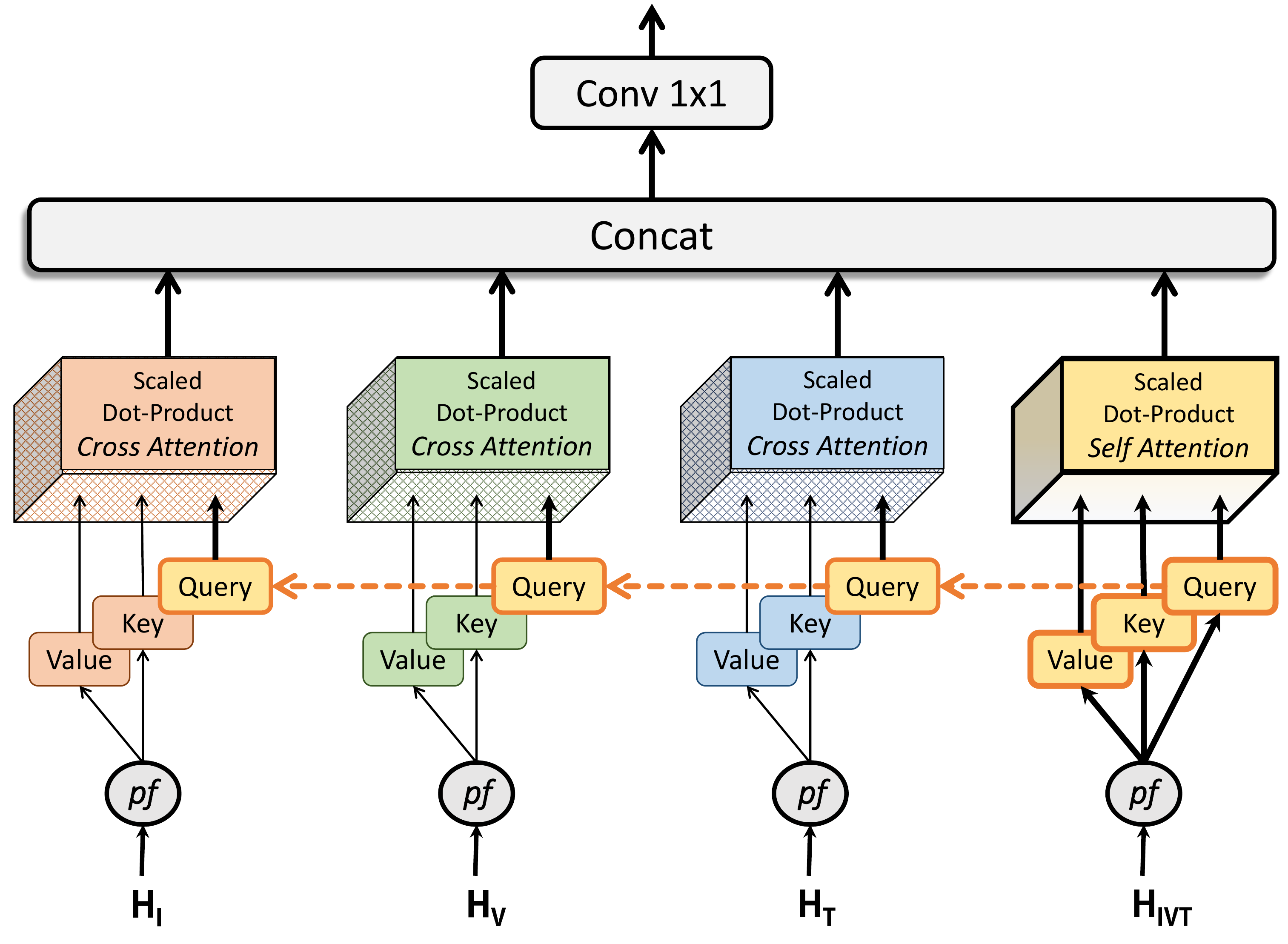}
\caption{Architecture of the multi-head of mixed attention (MHMA): {\normalfont showing the feature projection into Q, K and V, and subsequent multiple heads of self and cross attentions using scale-dot product attention mechanism}.
}
\label{fig:architecture:mhma}
\end{figure}

The multi-head attention combines both self- and cross-attentions, encouraging high-level learning of triplets from the interacting components as shown in Fig. \ref{fig:architecture:mhma}. 
It starts with a projection function, {\it \B{pf}}, which generates a set of value $\B{V}$, key $\B{K}$, and/or query $\B{Q}$ for each context feature ($\B{H}_I, \B{H}_V, \B{H}_T, \B{H}_{IVT}$).
In the implementation as shown in Equation \ref{eqn:methods:projection}, the {\it \B{pf}} function generates {vectors of $\B{Q}\in\mathbb{R}^{1\times C}$ and $\B{K}\in\mathbb{R}^{1\times C_\Zstroke}$} that represent the abridged mean descriptors of the contexts by leveraging the global average pooling (GAP) operation. 
Here, $C=100$ for triplet, whereas $C_\Zstroke=[6,10,15,100]$ for either instrument, verb, target, or triplet classes, respectively.
We map each descriptor to a feature embedding layer where we mask (dropout $\lambda=0.3$) parts of $\B{Q}$ to avoid repeating the same query in the $L$ alternating layers.
Using the {\it \B{pf}} function, we also obtain the $\B{V}\in\mathbb{R}^{H\times W\times C_\Zstroke}$ by a convolution operation on the feature context and reshape to $\mathbb{R}^{HW\times C_\Zstroke}$. 
{Hence, the extracted $\B{Q}$, $\B{K}$, and $\B{V}$ features follow the aforementioned decoding-by-attention concept (items 1 \& 2). 
The {\it \B{pf}} function generates each \B{K} and \B{Q} using FC layers as done in \citep{vaswani2017attention,dosovitskiy2020image}, and generates the \B{V} using convolution layers as done in \citep{fu2019dual,wang2018non,huang2019ccnet}.
}
\begin{equation}
    \B{\it pf}~(H) =
        \begin{cases}
            ~Q: & \quad FC~\Big(~DROPOUT~\Big(~GAP~\big(~{\bf H}~\big)~\Big)~\Big), \\[1.5mm]
            ~K: & \quad FC~\Big(~GAP~\big(~{\bf H}~\big)~\Big), \\[1.5mm]
            ~V: & \quad ~CONV~\big(~{\bf H}~\big).
        \end{cases}
    \label{eqn:methods:projection}
\end{equation}

Next, we build 4 attention heads for the instrument, verb, target, and triplet attention features.
In the existing Transformer and Transformer-based models, each of the heads learns a self-attention.
Self-attention helps a model understand the underlying meaning and patterns within its own feature representation. This is needed for initial scene understanding.
However, when each feature representation (such as a class-map) has been discriminated to attend to only one component in an image scene, understanding their underlying relationship requires a cross-attention across the component features.
In a cross-attention mechanism, the attention built from one context (the {\it source}) is used to highlight features in another context (the {\it sink}) as done in \citet{mohla2020fusatnet}.
While the self-attention mechanism computes the focal representation on the same triplet features, cross attentions learn the triplet representations by drawing attention from the individual components: namely instrument, verb, and target.
This models how the features of each component affect the triplet composition, by propagating the affinities from their respective context features to the required triplet features.

To utilize both self and cross attentions, we model the source context from the encoded class-map features ($\B{H}_I, \B{H}_V, \B{H}_T$) representing the triplet components and the sink context from the triplet features ($\B{H}_{IVT}$). Of course, the source context remains the same as the sink in the self-attention mechanism.
This means we generate the corresponding $\B{K}$s and $\B{V}$s from both the source and sink contexts, but generate the $\B{Q}$ only from the sink context using the projection function, {\it \B{pf}}, as shown in Fig. \ref{fig:architecture:mhma}. With $\B{Q}$ coming from the triplet features, we actually focus the image understanding on the actions of interest by pointing the cross-attention heads at the component's discriminative features ($\B{H}_{I}$, $\B{H}_{V}$, $\B{H}_{T}$) in a manner that helps the attention network benefit from the learnt class representations.
\begin{figure}[!t]
\centering
    \includegraphics[width=0.50\linewidth]{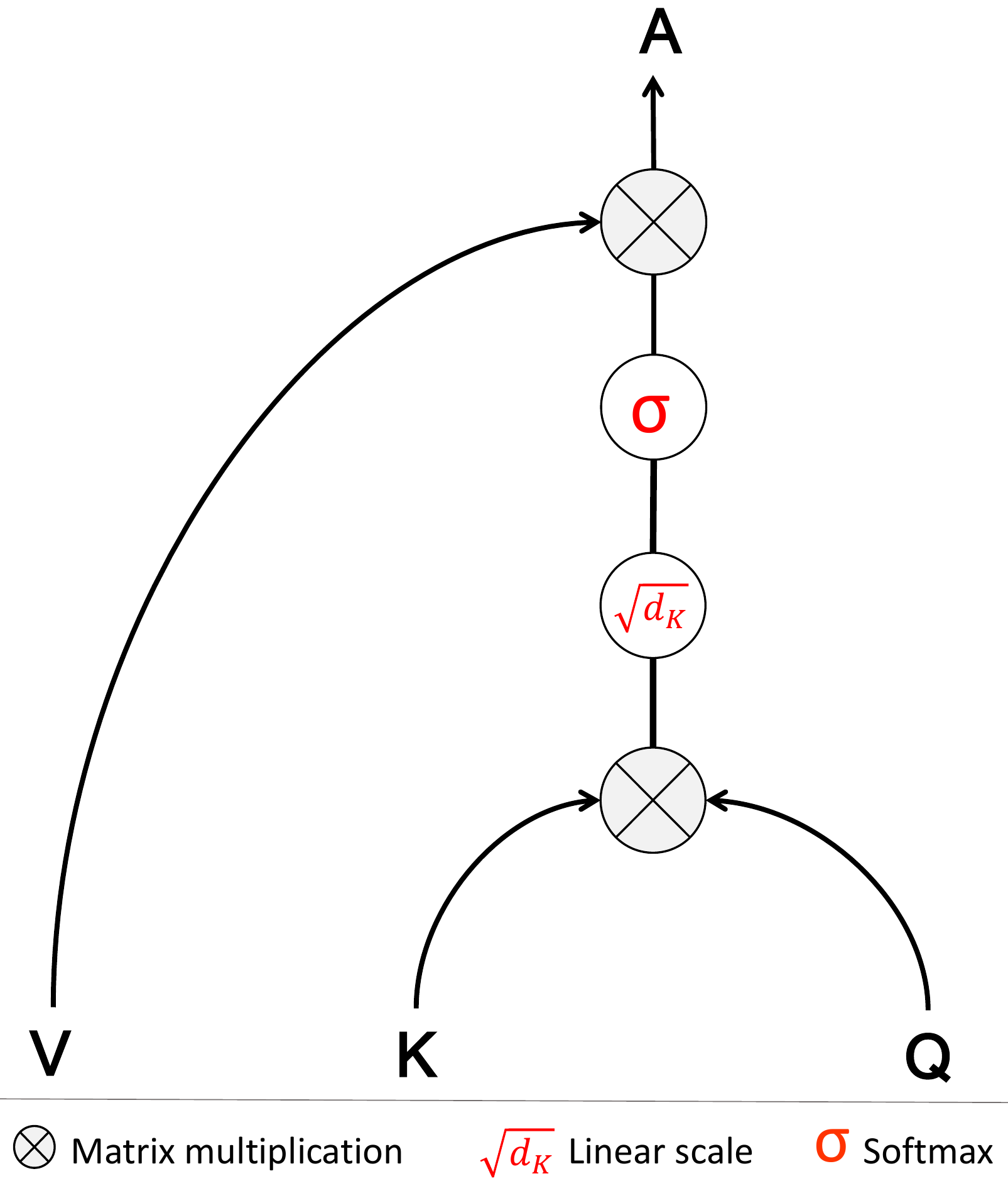}
\caption{Structure of scale dot-product attention mechanisms: \normalfont in self-attention, the (K,V,Q) triple comes from one feature context, whereas in cross-attention, the (K,V) pair comes from the source feature context while Q comes from the sink feature context.}
\label{fig:architecture:attention}
\end{figure}
This also respects the aforementioned decoding-by-attention concept.
We then learn a scaled dot product attention of the $\B{Q}$ on the ($\B{K,V}$) pair for each attention head as shown in Fig. \ref{fig:architecture:attention}.
Specifically, we derive the scaled dot product attention using the widely used attention formula {\citep{vaswani2017attention}} in Equation  \ref{eqn:methods:attn}:
\begin{equation}
    {\bf A}({\bf Q,K,V}) = {\bf V}.\sigma \left( \frac{{\bf KQ}^T}{\sqrt{d_{\bf K}}}\right),
    \label{eqn:methods:attn}
\end{equation}
where $\sigma$ is a softmax activation function, $\sqrt{d_{\bf K}}$ is a scaling factor, and $d_{\bf K}$ is the dimension of ${\bf K}$ after linear transformation.
The cross attention is implemented on the instrument, verb, and target attention heads, whereas self-attention is implemented on the triplet attention head.
While each attention head simultaneously concentrates on its own features of interest, the multi-head module combines heads ${\bf A}_{1..\mathcal{N}}$ to jointly capture the triplet features as in Equation \ref{eqn:ass}:
\begin{equation}
    {\bf A}_{1..\mathcal{N}} = {\bf W} \Big ( \mathbin\Vert^\mathcal{N}_{i=1} {\bf A}_i \Big) ,
    \label{eqn:ass}
\end{equation}
where $\mathbin\Vert$ is a concatenation operator for $\mathcal{N}=4$ attention heads. {\it {\bf A}$_1$} is the triplet self-attention, {\it {\bf A}$_{2 . . . \mathcal{N}}$} are the triplet cross attentions with the interacting components. ${\bf W}$ is the matrix of convolution weights. This packed convolution scheme merges the information from all attention heads while preserving its spatial structure.

\subsubsection{Feed-forward}
The output of the multi-head attention is further refined by a feed-forward layer which is a stack of 2 convolutions with an AddNorm.
The output is a refined $\B{H}_{IVT}$ with each channel attending to each triplet class.

\subsection{Triplet Classification} 
\label{sec:methodology:transformer:heads}
{The RDV model terminates with a linear classifier for the final classification of the triplets.}
In this layer, we apply a global pooling operation on the $\B{H}_{IVT}$ from the $L^{th}$ layer of the RDV decoder, followed by an FC-layer (with $C=100$ neurons) for the triplet classification.
The output logits ($\B{Y}_{IVT}$) are trained jointly end-to-end with the auxiliary logits from the encoder.

\subsection{Attention Tripnet}

In our previous work \citep{nwoye2020recognition}, Tripnet relies on two modules: (1) the class activation guide (CAG), which leverages instrument activations to detect verbs and targets via concatenated features, and (2) the 3D interaction space (3Dis), where features corresponding to the three components are projected in an attempt to resolve their interactions.

As an ablation model, we extend this to {\it Attention Tripnet} by only replacing the CAG in Tripnet \citep{nwoye2020recognition} with CAGAM as shown in Fig. \ref{fig:architecture:a-triplet}. This validates the contribution of attention modeling for verb and target detections using Attention Tripnet.

%% file: 05-experiment.tex
\section{Experiments}
\label{sec:exp}

\begin{figure*}[!ht]
\centering
    \includegraphics[width=0.7\textwidth]{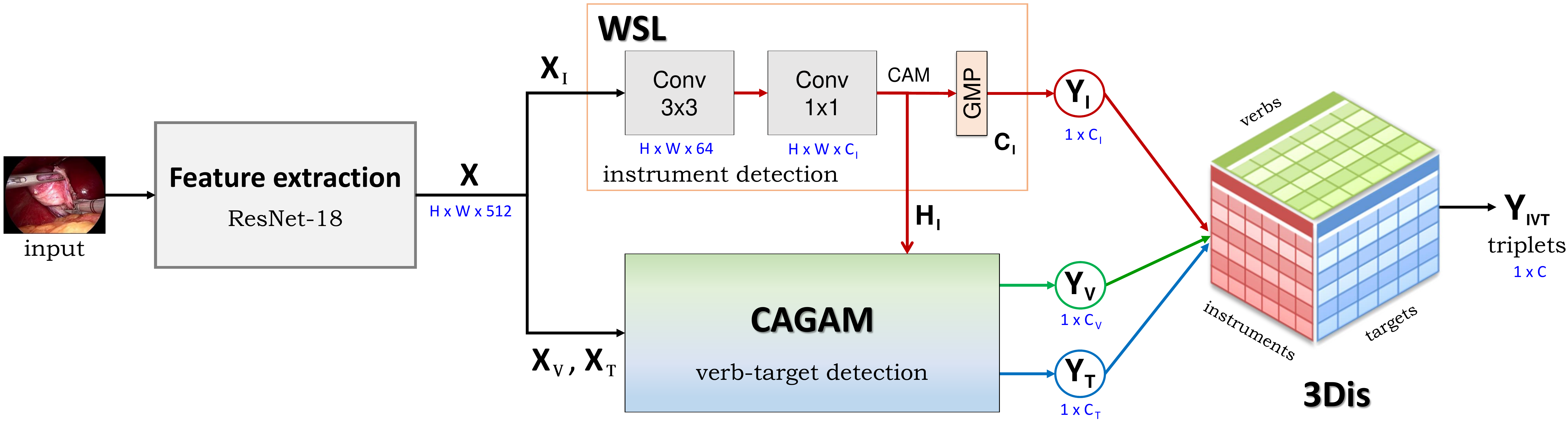}
\caption{Architecture of the Attention Tripnet showing the base (feature extraction backbone), neck (instrument detection branch and CAGAM module), and head (3D interaction space).
{\it Feature dimension values ($H=32, ~W=56, ~C_I=6, ~C_V=10, ~C_T=15, ~C=100$)}
}
\label{fig:architecture:a-triplet}
\end{figure*}

\subsection{Data Setup}
Due to variability in the video dataset, frame resolution varies from $480\times854$ to $1080\times1920$. We unified these spatial dimensions by resizing them to $256\times 448$. 
We also employed random scaling $[0.5,1.5]$ and brightness/contrast shift $(delta=0.2)$ as data augmentation for training.
The models are trained on 35 videos, validated on 5 videos, and tested on 10 videos according to the data split in Table \ref{table:data_split}.
To obtained specific labels for the component tasks, we design a mapping function, which extracts per-component labels from the triplet labels; those are three vectors of binary presence labels with length $N=[6,10,15]$ per frame, where $n\in N$ is the class size for each triplet's component trained as auxiliary task.
For a high-performance data loading pipeline, we store our training data as serialized TFRecords binaries.

\subsection{Training and Loss Functions}\label{sec:exp:training}
Since classifying each triplet component, namely instrument, verb, and target, is a multi-label classification problem, we employ weighted sigmoid cross-entropy losses: $L_I$, $L_V$, and $L_T$ respectively. The weighted cross-entropy with logits is as follows:
\begin{equation}
\begin{aligned}
L = \sum^C_{c=1}\frac{-1}{N}\Big(W_cy_clog\big(\sigma\left(\hat{y}_c\right)\big)  
    + \big(1-y_c\big)log\big(1-\sigma\left(\hat{y}_c\right)\big)\Big),
\end{aligned}
\end{equation}
where $y_c$ and $\hat{y}_c$ are respectively the ground truth and predicted labels for class $c$, $\sigma$ is the sigmoid function, and $W_c$ is a weight for class balancing.
The three component detection tasks are jointly learned in a multi-task manner following the uncertainty loss procedure given in \citet{kendall2018multi} that uses learnable parameters $w_I$, $w_V$, $w_T$ to automatically balance the tasks training as follows:
\begin{equation}
    L_{comp} = \frac{1}{3}\left(\frac{1}{e^{w_I}}L_I + \frac{1}{e^{w_V}}L_V + \frac{1}{e^{w_T}}L_T + w_I + w_V + w_T \right)
\end{equation}
This is only used for the auxiliary tasks captured by multi-task learning.

The triplet association loss $L_{assoc}$ is also modeled as a sigmoid cross-entropy.
To jointly learn the complete tasks end-to-end, we define the total loss ($L_{total}$) using the equation:
\begin{equation}
    L_{total} = L_{comp} + \rho L_{assoc} + \lambda L_2,
\end{equation}
where $\rho$ is a warm-up parameter that allows the network to focus solely on learning the individual components' information within the first 18 epochs. $\lambda = 1e^{-5}$ is a regularization weight decay for the $L_2$ normalization loss.

\subsection{Hyper-parameters}
The feature extraction backbone is pretrained on ImageNet.
All the models are trained using Stochastic Gradient Descent with Momentum $(\mu=0.95)$ as optimizer. We maintain a step-wise learning rate $(\eta=0.001)$ policy, decayed by $\delta=0.1$ after every $50$ epochs.
The models are trained in batches of size $8$ for $200$ epochs. The final model weights are selected based on their validation loss saturation. All the hyper-parameters are tuned on the validation set (5 videos) with up to 74 grid search experiments.

\subsection{Hardware and Schedule}
Our networks are implemented using TensorFlow and trained on GeForce GTX 1080 Ti, Tesla P40, RTX6000, and V100 GPUs. Full training takes approximately 118-180 hours on a single GTX 1080 Ti.
Total storage space consumption for the model, input data, output weights, and summaries is under 10GB.
Parameter counts for the MTL baseline, Tripnet, Attention Tripnet, and 8-layer RDV models reach 14.94M, 14.95M, 11.81M, 16.61M respectively.

\subsection{Inference and Evaluation Protocols}
\label{sec:results:metrics}
Model outputs are probability scores that can be thresholded to indicate class presence or absence.
We statistically evaluate the model's performance at recognizing surgical actions as a triplet using three metrics:
\begin{enumerate}
    \item \textbf{Component average precision:}
    \label{sec:results:metrics:comp}
    This measures the average precision (AP) of detecting the correct components of the triplet, as the area under the precision-recall curve per class. Using this, we measure the AP for instrument ($AP_I$), verb ($AP_V$), and target ($AP_T$) detections.
    To use these metrics for the naive models or for any model that predicts only the triplet labels $\B{Y}_{IVT}$, we decompose their predictions into the constituting components ($\B{Y}_I,\B{Y}_V,\B{Y}_T$) following Equation \ref{eqn:derived-ap}:
    \begin{equation}
        \label{eqn:derived-ap}
        \begin{split}
            Y_I &= [~\Scale[0.67]{MAX}\left(Y_{IVT}|I=i\right)~~~ \forall ~i \in \{0,1,.., C_1\}~],\\
            Y_V &= [~\Scale[0.67]{MAX}\left(Y_{IVT}|V=v\right)~~ \forall v \in \{0,1,..,C_2\}~],\\
            Y_T &= [~\Scale[0.67]{MAX}\left(Y_{IVT}|T=t\right)~~~ \forall t \in \{0,1,..,C_3\}~],\\
        \end{split}
    \end{equation}
    where $C_1,C_2$ and $C_3$ are the class sizes for the instrument, verb, and target components respectively. This directly translates to obtaining the probability of a given component class as the maximum probability value among all triplet labels having the same component class label in a given frame. For instance, the predicted probability of a \textit{grasper} instrument in a frame is the maximum probability of all triplet labels having \textit{grasper} as their instrument component label.
    The ground truth for these components is also derived in the same manner.
    
    \item \textbf{Triplet average precision:}
    \label{sec:results:metrics:assoc}
    This measures the AP of recognizing the tool-tissue interactions by observing elements of the triplet in conjunction. Using the same metrics as \citet{nwoye2020recognition}, we measure the APs for the instrument-verb ($AP_{IV}$), instrument-target ($AP_{IT}$), and instrument-verb-target ($AP_{IVT}$). During the AP computation, a prediction is registered as correct if all of the components of interest are correctly identified (e.g. instrument and verb for $AP_{IV}$). The main metric in this study is  $AP_{IVT}$, which evaluates the recognition of the complete triplets.
    
    \item \textbf{Top-N recognition performance:}
    \label{sec:results:metrics:error}
    Due to high similarities between triplets, we also measure the ability of a model to predict the exact triplets within its top $N$ confidence scores. For every given test sample $x_i$, a model made an error if the correct label $y_i$ does not appear in its top N confident predictions $\hat{y}_i$ for that sample. Using this setup, we measure the top-5, top-10, and top-20 accuracies for the triplet prediction. 
    We also show the top $10$ predicted triplet class labels and their AP scores for a more insightful analysis of the model's performance. 
    
\end{enumerate}

Video-specific AP scores are computed per category, across all frames of a given video. Averaging those APs over all videos gives us the mean AP (mAP), serving as our main metric.

%% file: 06-results.tex
\section{Results and Discussion}
\label{sec:results}
In this section, we rigorously validate individual components of the Attention Tripnet and Rendezvous (RDV) through careful ablation studies. We then provide a comparative analysis with baseline and state-of-the-art (SOTA) methods to show our methods' superiority.

\subsection{Quantitative Results}
\label{sec:results:quantitative}

\subsubsection{Ablation study on the encoder's attention type}

\input{tables/results-ablation-encoder-attention}

We begin with an ablation study for the choice of the attention type in the CAGAM module. We compare the module with a baseline model (MTL) \citep{nwoye2020recognition}, which implements a multi-task learning of instruments, verbs, and targets in separate branches with no attention (None), and show that attention guidance helps better detect the components in general (Table \ref{tab:results:qunatitative:ablation:attention}).
We also justify the distinct attention types for verbs and targets.
Firstly, the channel attention is used for both verb and target detections {\red(row 3)}, and the position attention is used for both verb and target detections (row 4), before they are combined (Dual attention) in the last row.
Channel attention is better suited for verbs than targets, with $+10.6\%$ vs $+3.5\%$ improvement respectively. Position attention behaves the opposite: $+2.8\%$ vs $+6.9\%$. Matching verbs with channel attention and targets with position attention gives the most balanced and highest improvement: $+12.4\%$ verbs, $+12.0\%$ targets. We, therefore, retain this choice in the proposed models.

\subsubsection{Ablation Study on Decoder's Attention Type}

\input{tables/results-ablation-decoder-attention}

One of the novel contributions of this work is its hybrid multi-head attention mechanism for resolving tool-tissue interactions, combining self- and cross-attention. This is a substantial innovation over transformers found in sequence modeling, which instead rely on multi-heads of self-attention only. Our choice of multi-head attention is justified in the following ablation study presented in Table \ref{tab:results:qunatitative:ablation:multihead}.

Our first ablation model in this regards ({\it Single Self}) uses a multi-head attention with the input feature coming from the high-level features ({\bf X}) of ResNet-18 to compute a successive scale dot-product attention over 8 decoder layers as in RDV.
It can be observed that using a multi-head of self-attention coming from a single source (triplet features) yields insufficient results for action triplet recognition.

The {\it Multiple Self} ablation model, as a "self-attention only" version of the RDV, uses self-attention in all four contexts: instrument, verb, target, and triplet. The RDV clearly performs the best in terms of association, justifying our use of cross-attention.

\subsubsection{Scalability Study on Multi-Head Layer Size}

\input{tables/results-ablation-scalability}

\input{tables/results-summary}

We carried out a scalability study to observe the performance of the RDV when increasing the number of multi-head layers while keeping track of the number of parameters and GPU requirements.
These results presented in Table \ref{tab:results:qunatitative:ablation:scale} show that the proposed model improves when scaled up, at the cost of increased computational requirements.
To balance performance and resource usage, we choose $L=8$ as default settings in all our experiments. An 8-layer RDV with $>25$ FPS processing speed can be used in real-time for OR assistance.

More ablation studies on the sequence modeling of the class-wise features, use of auxiliary classification loss, etc., are provided in the supplementary material.

\subsubsection{Component Detection and Association mAP}
For ease of reference, we summarize the overall performance of the experimented models on the considered metrics for both triplet component detection and the recognition of their interactions in Table \ref{tab:results:qunatitative:summary}. 
As a baseline, we design a CNN model that models the triplet recognition as a simple classification of $100$ distinct labels without taking any special reference to the constituting components.
The performance of this baseline model shows that it is not sufficient to naively classify the triplet IDs without considering the triplet components.
Even a temporal refinement of the naive CNN model outputs using a (TCN) \citep{lea2016temporal} is still sub-optimal.
Multi-task learning (MTL) of the triplet components helps the model gain some performance, but still scores low on triplet association.
The MTL outperforms the TCN here likely because a temporal refinement would not matter much if a Naive CNN does not capture significant representative features for triplet recognition.
The Tripnet model proposed in \citep{nwoye2020recognition} leverages the CAG to improve the MTL in the triplet components detection. It also improves the interaction recognition AP$_{IVT}$ by {$2.4\%$} using the 3Dis.

The Attention Tripnet uses the CAGAM to further improve the Tripnet's verb detection by $5.7\%$ and target detection by $5.3\%$. The Attention Tripnet is on par for instrument detection AP; this is likely due to the instrument detection being already saturated. The overall performance does increase, with indeed a $3.4\%$ improvement for triplet recognition.
The RDV, on the other hand, uses a multi-head attention decoder to further improve the association performance ($+9.7\%$ on instrument-verb, $+10.5\%$ on instrument-target). It improves the overall final triplet recognition by $9.9\%$ mAP$_{IVT}$ compared to the SOTA, tripling the improvement from the Attention Tripnet.
A breakdown of per-class detection of the triplet components and their association performance is presented in the supplementary material.

\subsubsection{Top-N Triplet Recognition Performance}

\input{tables/results-topN}

In our multi-class problem with 100 action triplet classes, getting a comprehensive view of a model's strength is difficult.
Here we focus on the top N predictions. As shown in Table \ref{tab:results:qunatitative:topN}, when considering the model's top 20 predictions, the model records an AP of $\approx95\%$. The model's confidence however decreases when considering more top predictions, suggesting how closely related most of the triplet classes could be.

\input{tables/results-top-triplets}

\subsubsection{Surgical Relevance of the Top Detected Triplets}

The result of the top 10 correctly detected triplets for the experimented models, presented in Table \ref{tab:results:qunatitative:topIVT}, reveals the individual strengths of the experimented models in recognizing the tool-tissue interaction. 
All triplets predicted in the top results are clinically sensible, with none of the more unexpected instrument-verb or instrument-target pairings.

Of importance, triplets with high surgical relevance in cholecystectomy procedure, i.e., \triplet{clipper, clips, cystic duct or artery} and \triplet{scissors, cut, cystic duct or artery}, which are critical for safety monitoring, are better detected by the RDV than the SOTA.
The proposed models learn to detect rare but clinically important uses of surgical instruments in their top 10 correctly predicted labels. 
This holds true for ambiguous instruments, like the {\it irrigator} that is mostly used to aspirate or irrigate but can as well be used to dissect in rare cases ( \triplet{irrigator, dissect, cystic-pedicle}). Another detected rare case include \triplet{bipolar, coagulate, blood-vessel}. This suggests that the models effectively learned the surgical semantics of instrument usage even with small examples of peculiar classes.

The triplet \triplet{grasper, grasp, specimen-bag} always appears in the top 2 even though its prevalence (6K) is not particularly high, compared to triplets such as \triplet{hook, dissect, gallbladder} (29K), \triplet{grasper, retract, liver} (13K), etc. This may be due to its consistent appearance in the workflow, usually towards the end; another factor could be the discernability of the {\it bag}.

For every triplet in the top-10 predictions of both the SOTA and the proposed models, the performance is usually higher in the proposed models. Remarkably, the entire top 10 for the RDV is recognized at an AP above 50\%.
Compared to SOTA, the proposed models show improvements in the more complex task of detecting complete triplets and instrument-target while showing comparable performance for the visually simpler instrument-verb detection task (see supplementary material and statistical analysis in Table \ref{tab:results:qunatitative:p-values}).

\begin{figure}[h]
\centering
    \includegraphics[width=0.99\linewidth]{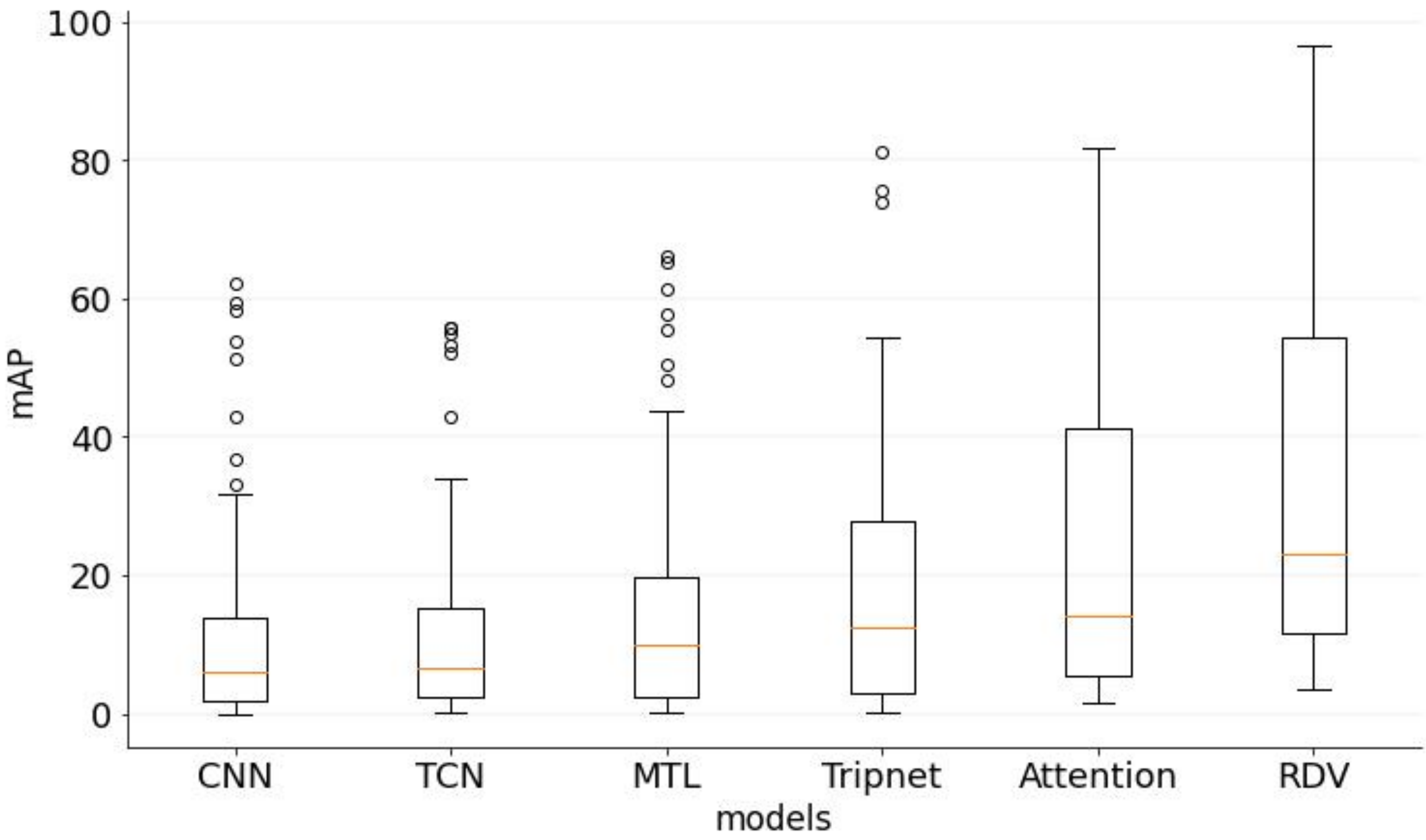}
    \vspace{-1mm}
    \caption{Distribution of the model AP for the  100 triplet class predictions. }
    \label{fig:results:quantitative:boxplot}
\end{figure}
In addition to the top 10, we also present the full extent of the model's performance on all 100 classes using the AP box plots in Fig. \ref{fig:results:quantitative:boxplot}, showing upper and lower performance bounds for each model as well the spread around the mean. 
The rectangular box represents the middle $50\%$ of the score for each model also known as {\it interquartile range}. As can be seen from Fig. \ref{fig:results:quantitative:boxplot}, the proposed models maintain higher median and upper-quartile performance than the baselines. They also maintain higher upper-whiskers showing the extent of their performance distribution above the interquartile range.

\subsubsection{Statistical Significance Analysis}

\input{tables/results-pvalues}

\begin{figure*}[t!]
\centering
    \includegraphics[width=1.0\linewidth]{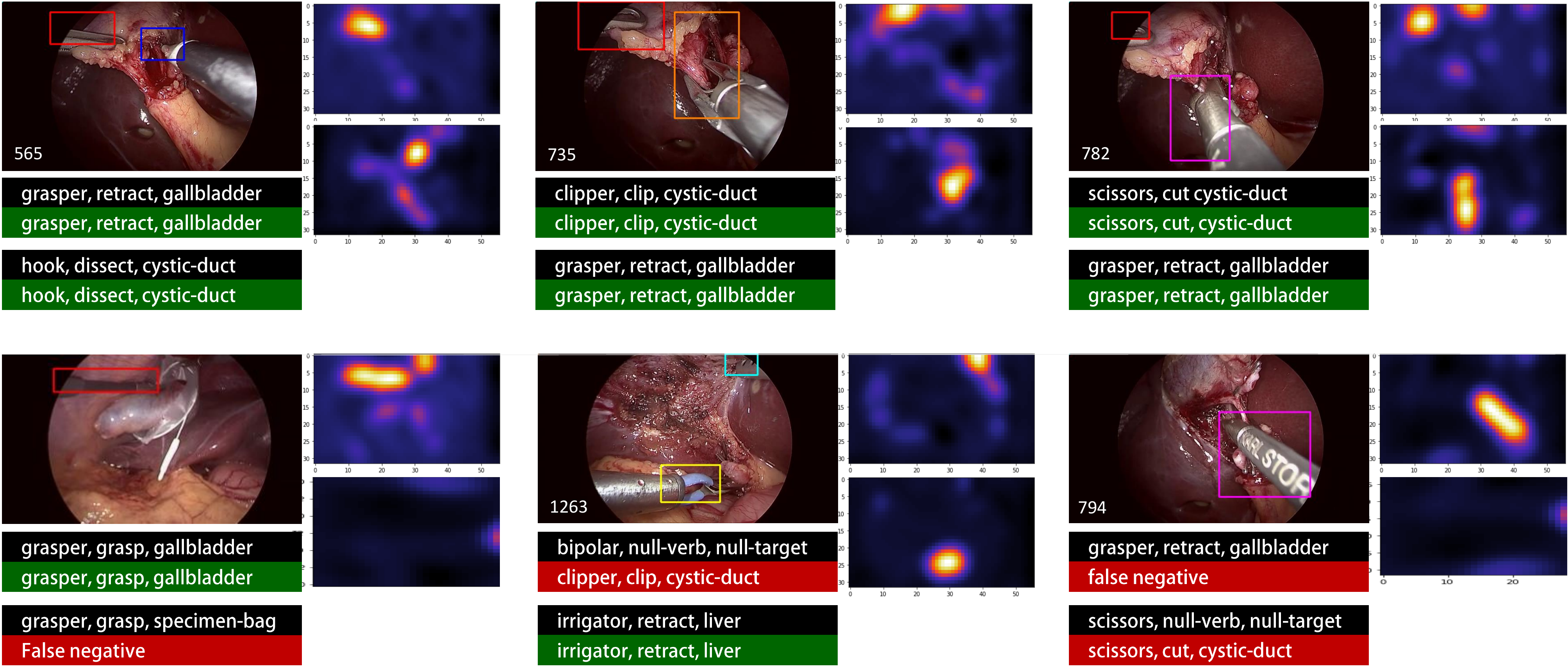}
\caption{Qualitative results showing the triplet predictions and the heatmaps for the triplet detection. Localization bounding boxes are obtained from the WSL module of the proposed RDV model. Predicted and ground truth triplets are displayed below each image: black = ground truth, green = correct prediction, red = incorrect prediction. A missed triplet is marked as false negative and a false detection is marked as false positive {\normalfont (Best viewed in color)}.
}
\label{fig:results:qualitative:loc}
\end{figure*}

We also measure the statistical significance of the proposed model performance using the SOTA model as the alternative method. Using the Wilcoxon signed-rank test, we sample $N = 30$ random batches of 100 consecutive frames instead of 30 random frames to simulate the evaluation on video clips.
The null hypothesis ($H_0$) states that the difference between the proposed method and the alternative method follows a symmetric distribution around zero.
We perform the significance analysis for each task, and based on the obtained $p$-values, presented in Table \ref{tab:results:qunatitative:p-values}, we draw the following conclusions:
\begin{enumerate}
    \item Both proposed models do not significantly improve the instrument detection sub-task. Their $p$-values fall short of the standard $0.05$. This is mainly because the instrument detection performance is already saturated in the alternative method; there is no new modeling in the proposed methods targeting their improvement. Being a two-tailed test, the $p$-value also shows that the SOTA does not outperform the proposed models on instrument detection.
    \item The guided attention mechanism is very useful in improving the verb and target detections in both the Attention Tripnet and RDV models. Their contributions are significant enough to even beat a more narrow $0.01$ significant level.
    \item Our contributions are also significant in improving the recognition of the tool-tissue interaction, with Attention Tripnet's improvement on $AP_{IVT}$ relevant at a $0.005$ significance level. Our best method (RDV) is more significant, with a $p$-value far below $0.001$.
\end{enumerate}
In summary, we reject the null hypothesis $H_0$ at a confidence level of 5\%, concluding that there is a significant difference between the proposed models and the alternative method.

\subsection{Qualitative Results}

\subsubsection{Triplet Recognition with Weak Localization}

The predicted class labels are obtained by applying a 0.5 threshold on the output probabilities of the proposed RDV model. We present those predicted labels in Fig. \ref{fig:results:qualitative:loc}, alongside the localization of the regions of action obtained from the weakly supervised learning (WSL) module of the network.
This localization, depicted by bounding boxes overlaid on the image, shows the focus of the model when it makes a prediction, thereby providing insight into its rationale. Those results are solid arguments in favor of the model's ability for spatial reasoning when recognizing surgical actions.
This suggests that the model can be further exploited for action triplet detection and segmentation.
We also provide a short {\it video} of this qualitative performance in the supplementary material (also accessible via:~~\videolink).

\subsubsection{Qualitative Analysis of Top 5 Predicted Triplets}
We also examine the top 5 prediction confidence of the proposed models compared to baselines on random frames (Fig. \ref{fig:results:qualitative:topN}). Fully correct predictions are signaled by the color green, while red indicates errors on all three components. Other colors indicate partially correct predictions.
RDV and Attention Tripnet outperform the baseline (MTL) and SOTA (Tripnet) each time, with the two actions correctly recognized each time within their top 5 predictions. 
Moreover, other actions in their top 5 have relevant components, showing these models' understanding of surgical actions by clustering triplets related to the performed actions.
More qualitative results on this are provided in the supplementary material.

\subsubsection{Attention Map Visualization}
To understand the benefit of the CAGAM's attention mechanism, we visualize its attention maps in Fig. \ref{fig:results:qualitative:attn}. For each input image, we randomly selected a few points ({\it marked} $i\in[1,2,3,4]$) in the images and reveal the corresponding attention maps for the tool-tissue interaction captured in the CAGAM's position attention map.
We observe that the attention module could capture semantic similarity and full-image dependencies, which change based on the contribution of the selected pixel to the action understanding.
This shows that the model learns attention maps that contextualize every pixel in the image feature in relation to the action performed.
For instance in the top image: point $2$, a pixel location on the instrument - {\it grasper}, creates an attention map that highlights both the instrument and its target - {\it gallbladder}.
Indeed, the attention guidance introduced in this model helps to highlight the triplet's interest regions while suppressing the rest.
This effect is shown further in the supplementary video.

%% file: tables/results-ablation-encoder-attention.tex
\begin{table}[h]
    \centering
    \setlength{\tabcolsep}{13pt}
    \caption{Ablation study on the task-attention suitability}
    \label{tab:results:qunatitative:ablation:attention}
    \resizebox{\columnwidth}{!}{%
    \begin{tabular}{@{}llccc@{}}
    \toprule
    Guided detection & & $AP_{V}$ & & $AP_{T}$ \\
    \midrule
    None ({\it as in} MTL baseline) && 48.4 && 28.2 \\
    CAM ({\it as in} Tripnet's CAG) && 51.3 && 32.1 \\
    CAM + Channel attention && 59.0 && 31.5 \\
    CAM + Position attention && 51.2 && 35.1 \\
    CAM + Dual$^1$  attention && \bf 61.1 && \bf 40.2 \\ 
    \bottomrule
    \end{tabular}
    }
    \footnotesize{\\$^1$ Dual = (channel + position)~ attentions}
\end{table}

%% file: tables/results-ablation-decoder-attention.tex
\begin{table}[h]
\centering
    \setlength{\tabcolsep}{6pt}
    \caption{Ablation study on the attention type in the multi-head decoder}
    \label{tab:results:qunatitative:ablation:multihead}
    \resizebox{\columnwidth}{!}{%
    \begin{tabular}
        {@{}lrccccc@{}}\toprule
        Model && Layer size & AP$_{IV}$ & AP$_{IT}$ & AP$_{IVT}$ \\ \midrule
        Single Self && 6 &  29.8 & 23.3 & 18.8  \\ 
        Multiple Self  && 6 &  35.7 & 32.8 & 26.1 \\
        Self + Cross (RDV)  && 6 &  \bf 39.4 & \bf 36.9 & \bf 29.9 \\
        \bottomrule
    \end{tabular}
 }
\end{table}

%% file: tables/results-ablation-scalability.tex
\begin{table}[h]
\centering
    \setlength{\tabcolsep}{4pt}
    \caption{A scalability study on the multi-head layer size: {\normalfont showing the mean average precision (mAP) for varying triplet associations, number of learning parameters (Params) in millions (M), and inference time (i-Time) in frame per seconds (FPS) on GTX 1080 Ti GPU.}}
    \label{tab:results:qunatitative:ablation:scale}
    \resizebox{\columnwidth}{!}{%
    \begin{tabular}
    {@{}lcccccc@{}}\toprule
    Layer size & \phantom{abc} & \makecell{$mAP_{IV}$\\(\%)$\uparrow$} & \makecell{$mAP_{IT}$\\(\%)$\uparrow$}& \makecell{$mAP_{IVT}$\\(\%)$\uparrow$ \tnote{1}} & \makecell{Params \\(M)$\downarrow$} & \makecell{i-Time \\(FPS)$\uparrow$}  \\ \midrule
    1  && 35.8 & 30.7 & 24.6 & 12.6 & 54.2\\
    2  && 36.0 & 41.1 & 27.0 & 13.1  & 47.9\\
    4  && 38.5 & 32.9 & 27.3 & 14.3 & 39.2\\
    8  && 39.4 & 36.9 & 29.9 & 16.6 & 28.1\\
    \bottomrule
    \end{tabular}
    }
\end{table}

%% file: tables/results-summary.tex
\begin{table*}[!t]
\centering
    \setlength{\tabcolsep}{7pt}
    \caption{Performance summary of the proposed models compared to state-of-the-art and baseline models}
    \label{tab:results:qunatitative:summary}
    \resizebox{\textwidth}{!}{%
    \begin{tabular}{@{}llcclllllcclllll@{}}\toprule
    \multicolumn{2}{c}{\multirow{2}{*}{Method}}&\phantom{abc}&&
    \multicolumn{5}{c}{Component detection}&\phantom{abc}&&
    \multicolumn{5}{c}{Triplet association}\\ \cmidrule{5-9} \cmidrule{12-16} 
    \multicolumn{2}{l}{} &&& $AP_{I}$ && $AP_{V}$ && $AP_{T}$ &&& $AP_{IV}$ && $AP_{IT}$ && $AP_{IVT}$ \\ \midrule
    \multirow{3}{*}{\makecell[l]{Naive Baseline} }
        & CNN     &&&  57.7 && 39.2 && 28.3 &&& 21.7 && 18.0 && 13.6 \\ 
        & TCN     &&&  48.9 && 29.4 && 21.4 &&& 17.7 && 15.5 && 12.4  \\ 
        & MTL     &&&  84.5 && 48.4 && 28.2 &&& 26.6 && 21.2 && 17.6 \\\midrule
    SOTA & Tripnet \citep{nwoye2020recognition} &&&  {\bf 92.1} && 54.5 && 33.2 &&& 29.7 && 26.4 && 20.0  \\ \midrule
    \multirow{2}{*}{Ours}
      & Attention Tripnet &&& 92.0 &&  60.2 && {\bf 38.5} &&& 31.1 && 29.8 && 23.4  \\ 
      & Rendezvous &&& 92.0 && {\bf 60.7} && 38.3 &&& {\bf 39.4} && {\bf 36.9} && {\bf 29.9}  \\
        \bottomrule 
    \end{tabular}
    }
    
\end{table*}

%% file: tables/results-topN.tex
\begin{table}[h]
\setlength{\tabcolsep}{10pt}
\centering
    \caption{Top N Accuracy of the triplet predictions}
    \label{tab:results:qunatitative:topN}
    \resizebox{\columnwidth}{!}{%
    \begin{tabular}{@{}llrrr@{}}
        \toprule
        \multicolumn{2}{c}{Method} & 
        Top-5 & Top-10 & Top-20  \\\midrule
    \multirow{3}{*}{\makecell[l]{Naive\\Baseline} }
        & CNN &  67.0 & 80.0 & 90.2  \\ 
        & TCN    & 54.5 & 69.4 & 84.3 \\ 
        & MTL  & 70.2 & 80.2 & 89.5  \\\midrule
    SOTA & Tripnet & 70.5 & 81.9 & 91.4  \\ \midrule
    \multirow{2}{*}{Ours}
       & Attention Tripnet & 75.3 & 86.0 & 93.8 \\ 
       & Rendezvous & \bf 76.3 & \bf 88.7 & \bf 95.9  \\ 
       \bottomrule
    \end{tabular}
    }
\end{table}

%% file: tables/results-top-triplets.tex
\begin{table*}[t]
\centering
    \setlength{\tabcolsep}{6pt}
    \caption{Top-10 predicted Triplets (AP$_{IVT}$ for Instrument-Verb-Target Interactions).}
    \label{tab:results:qunatitative:topIVT}
    \resizebox{1.0\textwidth}{!}{%
    \begin{tabular}{@{}lrclrclr@{}}\toprule
        \multicolumn{2}{c}{Tripnet (SOTA)} & \phantom{abc} & \multicolumn{2}{c}{Attention Tripnet} & \phantom{abc} & \multicolumn{2}{c}{Rendezvous} \\
        \cmidrule{1-2} \cmidrule{4-5} \cmidrule{7-8}
        {Triplet} & {AP} && {Triplet} & {AP} && {Triplet} & {AP} \\ \midrule 
        grasper,retract,gallbladder  & 77.30  &   & grasper,grasp,specimen-bag  & 82.34  &   & grasper,retract,gallbladder  & 85.34 \\
        grasper,grasp,specimen-bag  & 76.50  &   & grasper,retract,gallbladder  & 78.41  &   & grasper,grasp,specimen-bag  & 81.75 \\
        bipolar,coagulate,liver  & 67.39  &   & bipolar,coagulate,liver  & 68.85  &   & hook,dissect,gallbladder  & 75.90 \\
        hook,dissect,gallbladder  & 57.54  &   & irrigator,dissect,cystic-pedicle  & 66.21  &   & grasper,retract,liver  & 66.70 \\
        irrigator,aspirate,fluid  & 57.51  &   & hook,dissect,gallbladder  & 63.22  &   & bipolar,coagulate,liver  & 63.12 \\
        grasper,retract,liver  & 54.25  &   & grasper,retract,liver  & 58.06  &   & clipper,clip,cystic-duct  & 59.68 \\
        clipper,clip,cystic-artery  & 47.44  &   & grasper,grasp,cystic-pedicle  & 55.35  &   & bipolar,coagulate,blood-vessel  & 57.18 \\
        scissors,cut,cystic-duct  & 42.57  &   & scissors,cut,cystic-artery  & 48.44  &   & scissors,cut,cystic-artery  & 53.84 \\
        scissors,cut,cystic-artery  & 40.37  &   & irrigator,aspirate,fluid  & 47.11  &   & irrigator,aspirate,fluid  & 51.95 \\
        clipper,clip,cystic-duct  & 39.62  &   & bipolar,coagulate,abdominal-wall-cavity  & 46.07  &   & clipper,clip,cystic-artery  & 51.52 \\
        \midrule
        mean & 56.05  &&& 61.41  &&& 64.70 \\
        \bottomrule
    \end{tabular}
    }
\end{table*}

%% file: tables/results-pvalues.tex
\begin{table}[h]
\centering
    \setlength{\tabcolsep}{11pt}
    \caption{The $p$-values obtained in Wilcoxon signed-rank test of the proposed methods using the SOTA model (Tripnet) as the alternative method. ({\it Lower $p$-value is preferred})}
    \label{tab:results:qunatitative:p-values}
    \resizebox{1.0\columnwidth}{!}{%
    \begin{tabular}{@{}llcc@{}}\toprule
        \multicolumn{2}{l}{\multirow{2}{*}{Tasks}} & \multicolumn{2}{c}{{ Proposed methods }} \\ \cmidrule{3-4} 
        \multicolumn{2}{c}{} & Attention Tripnet & Rendezvous \\ \midrule
        \multirow{3}{*}{\makecell[l]{Component \\Detection}} & $AP_{I}$ & $p\approx0.327$ & $p\approx0.374$ \\
            & $AP_{V}$ & $p\ll0.001$ & $p\approx0.003$ \\ 
            & $AP_{T}$ & $p\ll0.001$ & $p\ll0.001$ \\ \midrule
        \multirow{3}{*}{\makecell[l]{Triplet \\Association}} & $AP_{IV}$ & $p\approx0.018$ & $p<0.001$ \\ 
            & $AP_{IT}$ & $p\approx0.010$ & $p\approx0.005$ \\ 
            & $AP_{IVT}$ & $p<0.005$ & $p\ll0.001$ \\
        \bottomrule
    \end{tabular}
    }
\end{table}

%% file: 07-conclusion.tex
\section{Conclusion}
\label{sec:conclusion}

\begin{figure*}[!t]
\centering
    \includegraphics[width=1.0\linewidth, height=0.6\textwidth]{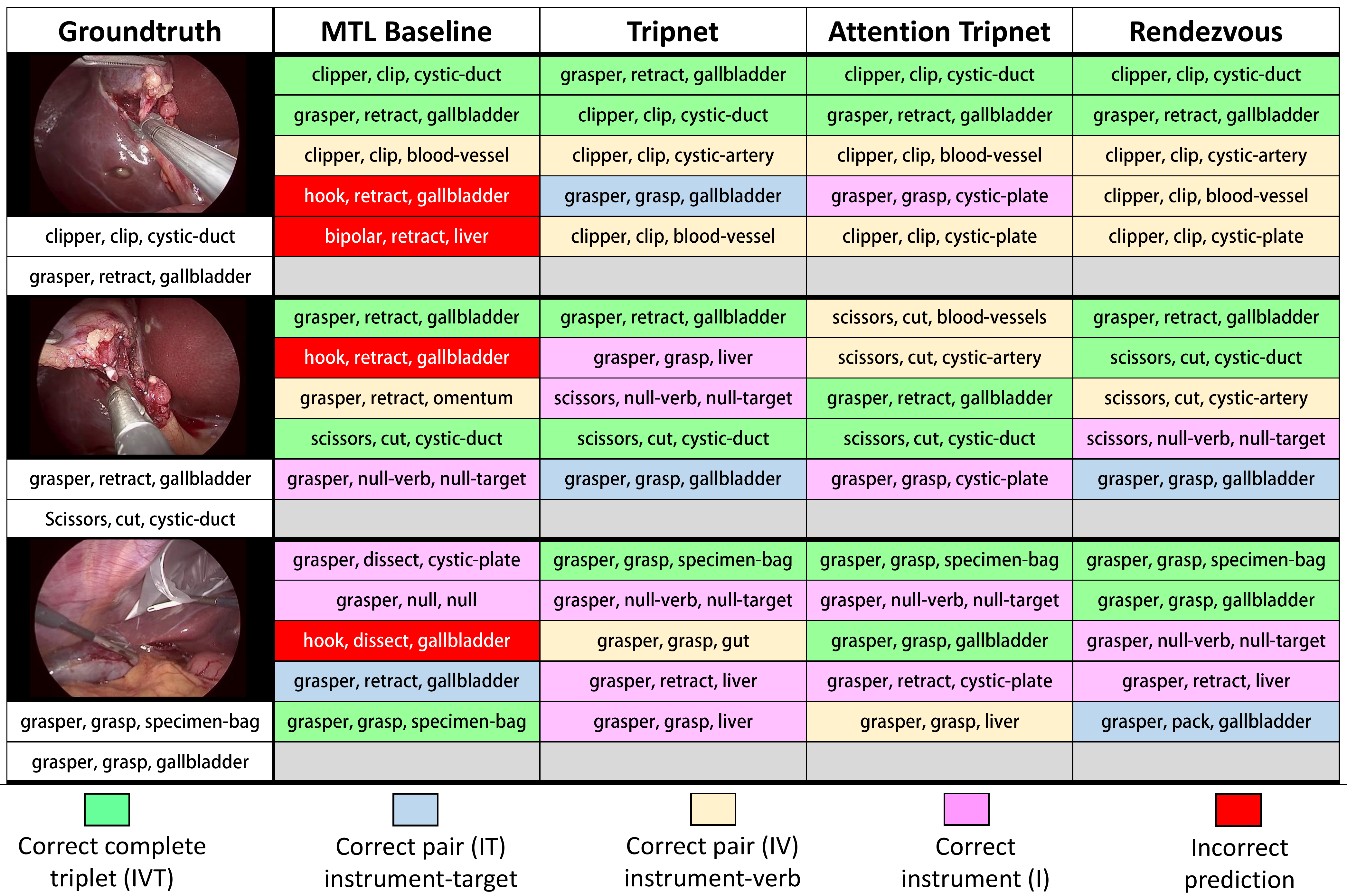}
\caption{Qualitative results showing the top-5 triplet predictions for the best performing baseline, SOTA, and the proposed models {\normalfont (Best viewed in colour)}.}
\label{fig:results:qualitative:topN}
\end{figure*}

\begin{figure*}[h!]
\centering
    \includegraphics[width=1.0\linewidth]{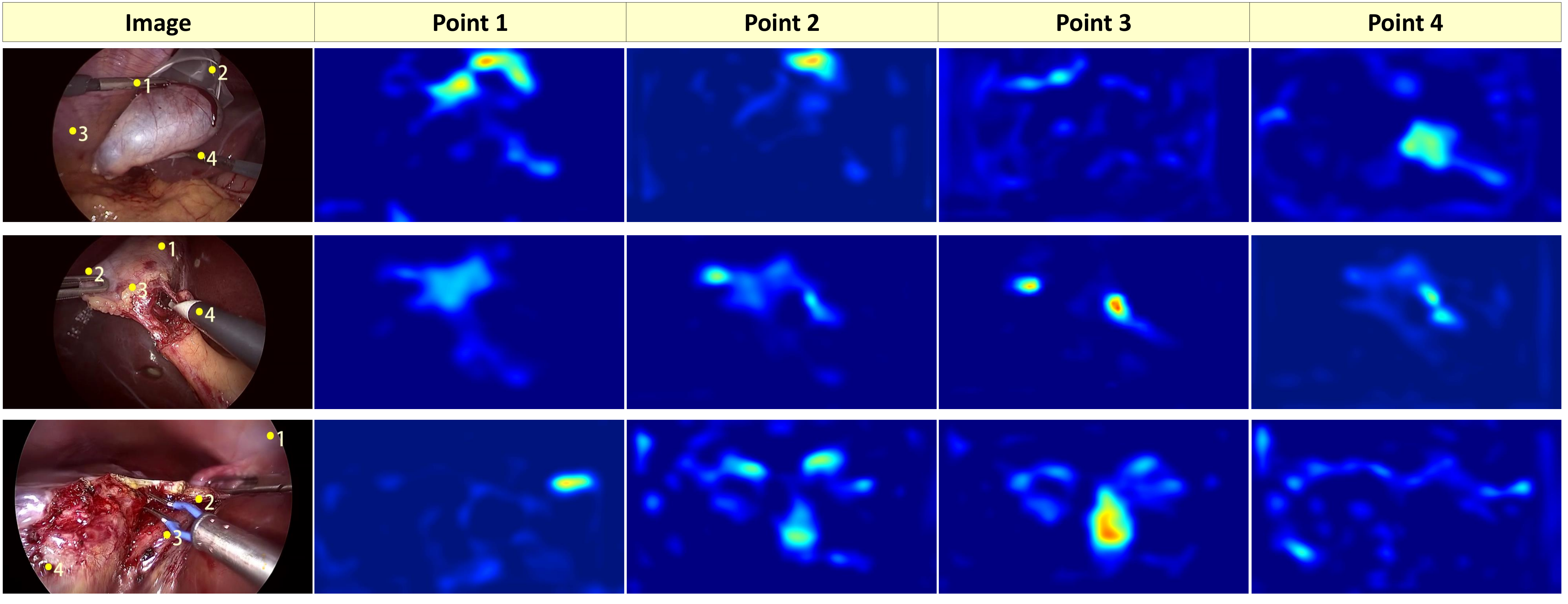}
\caption{Attention maps in the CAGAM module on the CholecT50 test set. The left column is the input image, the subsequent columns are the attention maps captured by the different points as marked in the input image. The attention map shows the focus on the target \normalfont{(best seen in color)}.
}
\label{fig:results:qualitative:attn}
\end{figure*}

We have presented methods featuring new forms of attention mechanisms that surpass the state-of-the-art for surgical actions triplet \triplet{instrument, verb, target} recognition. 
We first proposed a novel approach for attention intended for verbs and targets, using instrument class activation maps. We have also introduced a novel hybrid attention mechanism that resolves component association in triplets by leveraging multiple heads of both self and cross attentions on the component features.

We have rigorously validated our performance claims on {\it CholecT50}, a new large-scale endoscopic video dataset also contributed in this work.
We also discussed the benefits of the proposed methods in terms of clinical significance. Qualitative results suggest possible extensions to different tasks, including automated surgical report generation and spatial action segmentation.

While these initial results are encouraging, many challenges remain. One is the scalability on unseen triplets which may likely be tackled by zero-, one- or few-shot learning. Our results on rare triplets already hint at promising prospects for this approach. Inference speed is another challenge: increasing the number of layers generally drives up the performance, but is computationally very costly. Implementing a more lightweight Rendezvous would help alleviate some of these costs.

One limitation of this work is that target localization using the same weakly-supervised technique as for instruments is not yet achieved. This is likely due to the target's visibility not being the sole indicator for a positive binary label, both in the ground truth annotations and the model predictions.
{\red We also observed that it is not possible to recognize multiple instances of the same triplet, e.g., {\it two \triplet{grasper, grasp, gallbladder}}. This is due to the nature of the binary presence annotation, which does not provide an instance count for each unique triplet class. Only actions performed with the {\it grasper} instrument can have multi-instance occurrence in this dataset. Nonetheless, this does not affect recognition but is considered a limitation in future work on triplet localization, where multiple instances would need to be detected differently. }

With high-profile potential applications such as safety monitoring, skill evaluation, and objective reporting, our surgical action triplet method, as well as the release of our dataset for the 2021 Endoscopic Vision challenge, bring considerable value to the field of surgical activity understanding.

Future work will consider temporal modeling as some of the action verbs could be better recognized by the temporal dynamics of the tool-tissue interaction.

%% file: 08-acknowledgement.tex
\section*{Acknowledgements}
\label{sec:acknowledgement}
This work was supported by French state funds managed within the Investissements d\textsc{\char13}Avenir program by BPI France (project CONDOR) and by the ANR under references ANR-11- LABX-0004 (Labex CAMI), ANR-16-CE33-0009 (DeepSurg), ANR-10-IAHU-02 (IHU Strasbourg) and ANR-20-CHIA-0029-01 (National AI Chair AI4ORSafety). 
It was granted access to  HPC resources of Unistra Mesocentre and GENCI-IDRIS (Grant 2021-AD011011638R1).
The authors also thank the IHU and IRCAD research teams for their help with the data annotation during the CONDOR project.

%% file: 10-appendix.tex
\clearpage
\newpage
\renewcommand{\thesubsection}{\Alph{subsection}}
\appendix
\clearpage
\onecolumn
\pagenumbering{gobble}

\section*{\centerline{\large Rendezvous: attention mechanisms for the recognition of surgical action triplets in endoscopic videos}}
\subsection*{\centerline{===== ~~~Supplementary Material~~~ =====}}
\begin{center}
    \rule{\linewidth}{0.88pt}
\end{center}

\section{Additional Dataset Statistics}

\input{tables/stats-iv}
\input{tables/stats-it}

\section{More Quantitative Results}

\input{tables/results-perclass-i}
\input{tables/results-perclass-v}
\input{tables/results-perclass-t}
\input{tables/results-perclass-iv}
\input{tables/results-perclass-it}
\input{tables/results-ablation-sequence}
\input{tables/results-ablation-auxilliary}

\section{More Qualitative Results}

\begin{figure*}[!h]
\centering
    \includegraphics[width=0.99\linewidth]{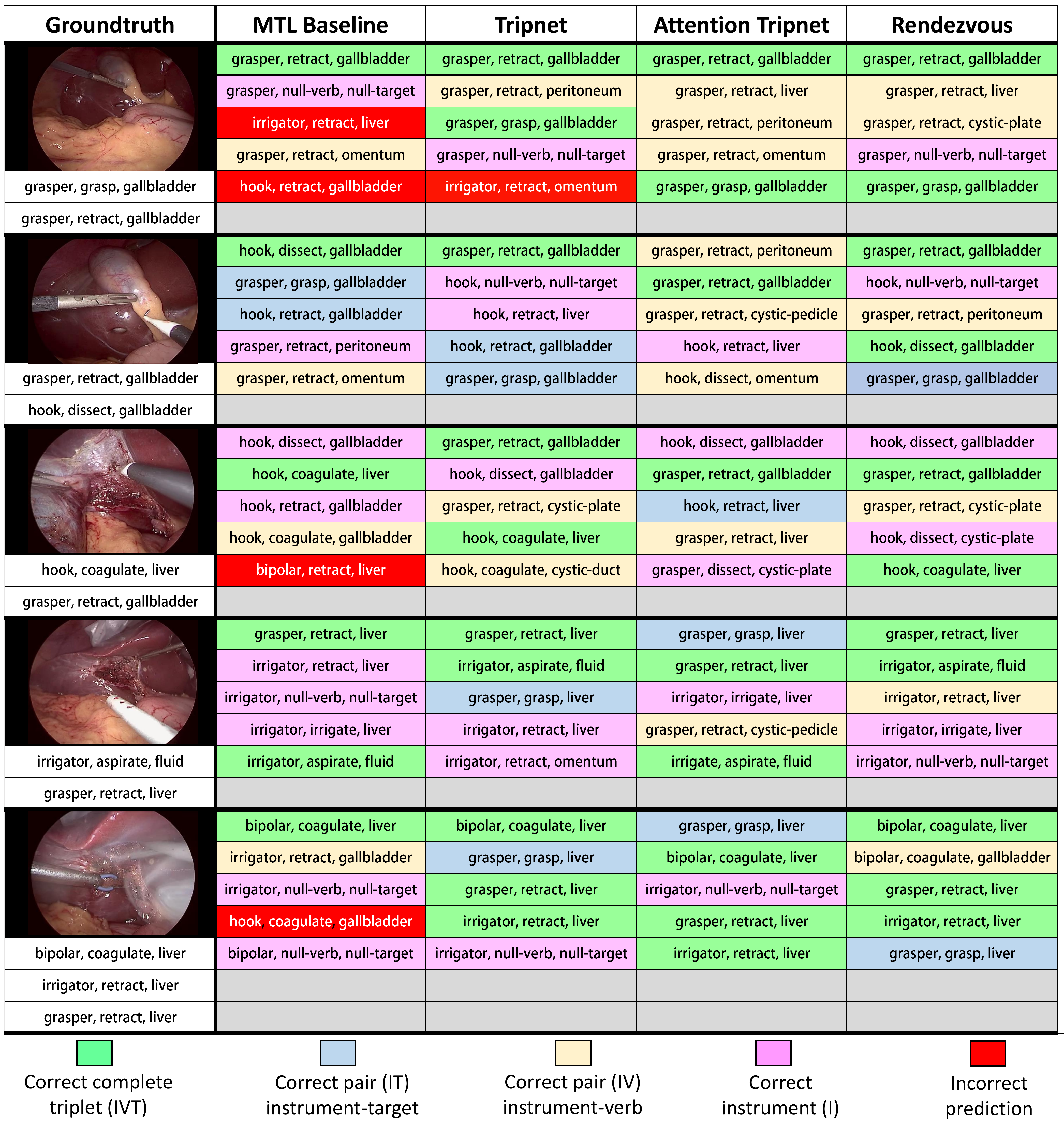}
\caption{More qualitative results on the top-5 triplet prediction for the best performing baseline, SOTA, and the proposed models ({\it Best view in colour.}).}
\label{fig:results:qualitative:topN-supp}
\end{figure*}

\section{Supplementary Video}

\begin{figure*}[!h]
  \begin{minipage}[c]{0.1\textwidth}
    \href{https://youtu.be/d_yHdJtCa98}{\includegraphics[width=0.8\linewidth]{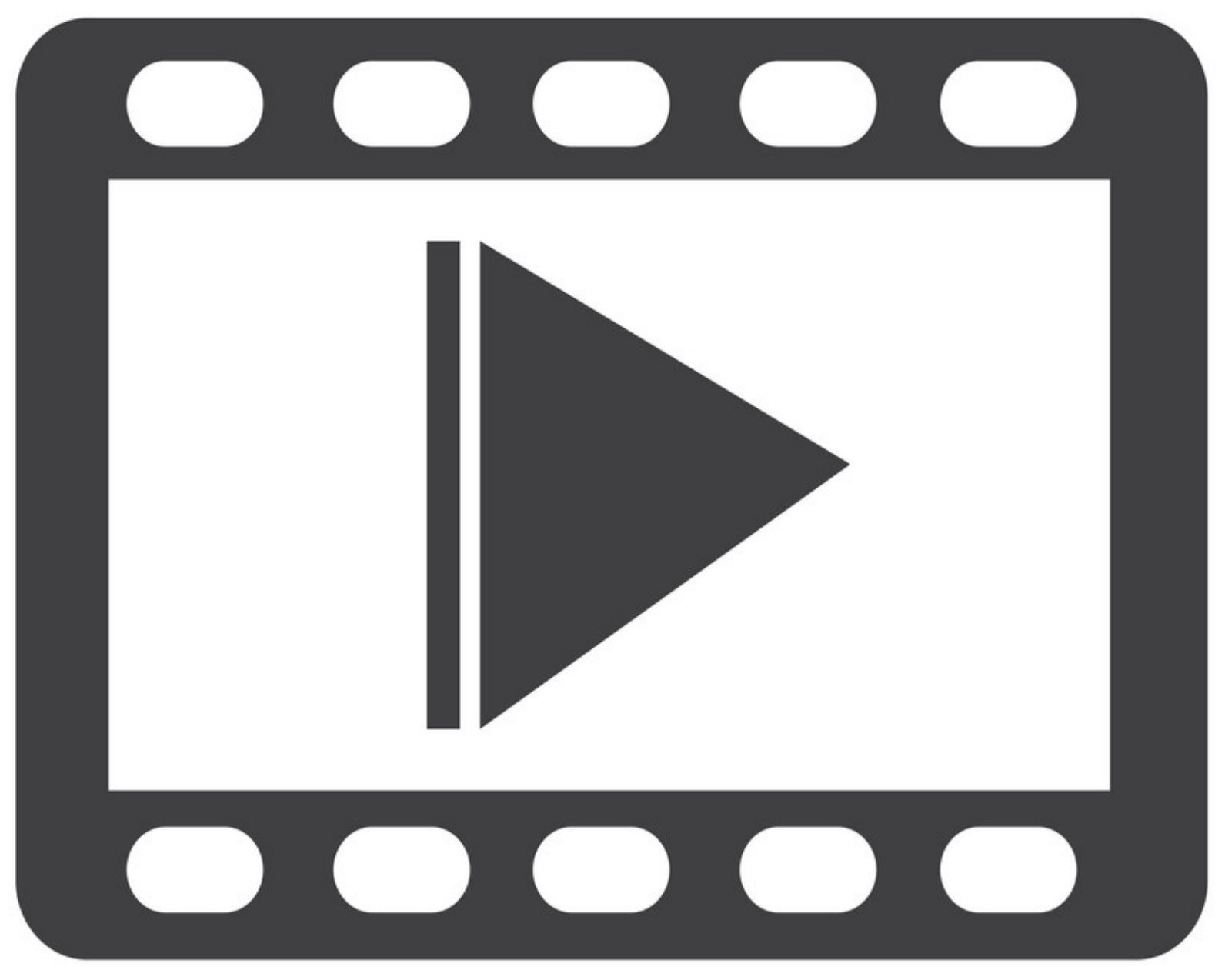}}
  \end{minipage}\hfill
  \begin{minipage}[c]{0.9\textwidth}
    {~\\Vid. D.1: A video showing some qualitative results of the RDV model for triplet prediction, action region localization, and attention maps is included as supplementary material, accessible online via~~ \videolink 
    } \label{fig:me}
  \end{minipage}
\end{figure*}

%% file: tables/stats-iv.tex
\begin{table*}[!h]
\setlength{\tabcolsep}{10pt}
\centering
\caption{Dataset statistics showing the instrument-verb co-occurrence distribution. }
\label{table:data_IV-supp}
\resizebox{\columnwidth}{!}{%
\begin{tabular}{@{}lcrrrrrrrrrr@{}}\toprule
\multirow{2}[4]{*}{Instrument}&\phantom{abc}&\multicolumn{10}{c}{Verb}\\\cmidrule{3-12}
    &  & grasp & retract & dissect & coagulate & clip & cut & aspirate & irrigate & pack & null\\ \midrule 
    grasper &  & 15743 & 69386 & 753 & - & - & - & - & - & 328 & 4759\\
    bipolar &  & 188 & 282 & 1026 & 4569 & - & - & - & - & - & 632\\
    hook &  & - & 658 & 47042 & 616 & - & 107 & - & - & - & 4397\\
    scissors &  & - & - & 157 & 17 & - & 1790 & - & - & - & 171\\
    clipper &  & - & - & - & - & 3070 & - & - & - & - & 309\\
    irrigator  &  & - & 469 & 269 & - & - & - & 3122 & 572 & - & 573\\
\bottomrule
\end{tabular}
}
\end{table*}

%% file: tables/stats-it.tex
\begin{table*}[!h]
\centering
\caption{Dataset statistics showing the instrument-target co-occurrence distribution. {\normalfont Target ids 0 ... 14 correspond to \textit{gallbladder, cystic-plate, cystic-duct, cystic-artery, cystic-pedicle, blood-vessel, fluid, abdominal-wall/cavity, liver, adhesion, omentum, peritoneum, gut, specimen-bag, and null respectively.}}}
\label{table:data_IT-supp}
\resizebox{\columnwidth}{!}{%
\begin{tabular}{@{}lcrrrrrrrrrrrrrrr@{}} \toprule
\multirow{2}[4]{*}{Instrument}&\phantom{abc}&\multicolumn{15}{c}{Target}\\ \cmidrule{3-17}
    &  & 0 & 1 & 2 & 3 & 4 & 5 & 6 & 7 & 8 & 9 & 10 & 11 & 12 & 13 & 14 \\\midrule 
    grasper &  & 56981 & 1446 & 1029 & 76 & 67 & - & - & - & 13729 & - & 4660 & 669 & 719 & 6834 & 4759\\
    bipolar &  & 728 & 472 & 247 & 255 & 86 & 251 & - & 434 & 2854 & 73 & 507 & 73 & - & 85 & 632\\
    hook &  & 29988 & 2907 & 7902 & 2994 & 15 & 93 & - & - & 368 & - & 3727 & 429 & - & - & 4397\\
    scissors &  & 52 & 32 & 808 & 613 & - & 21 & - & - & 90 & 155 & 137 & 56 & - & - & 171\\
    clipper &  & - & 53 & 1856 & 1097 & 13 & 51 & - & - & - & - & - & - & - & - & 309\\
    irrigator &  & 59 & 10 & 41 & - & 118 & - & 3122 & 413 & 480 & - & 189 & - & - & - & 573\\
\bottomrule
\end{tabular}
}
\end{table*}

%% file: tables/results-perclass-i.tex
\begin{table*}[!h]
\centering
\setlength{\tabcolsep}{13pt}
    \caption{Result Breakdown for Per-Class Instrument ($AP_I$) Detection: {\normalfont The proposed models correctly detect all the various categories of the surgical instrument at a performance higher than $80.0\%$}}
    \label{tab:results:qunatitative:instrument-supp}
    \resizebox{1.0\columnwidth}{!}{%
    \begin{tabular}{@{}llccccccr@{}}
    \toprule
    \multicolumn{2}{c}{Method} & Grasper & Bipolar & Hook & Scissors & Clipper & Irrigator & mAP \\ \midrule
        \multirow{3}{*}{\makecell[l]{Naive \\Baseline} }
	& CNN   & 91.4 & 47.9 & 89.1 & 24.0 & 50.2 & 43.2 & 57.7  \\ 
	& TCN   & 90.5 & 37.6 & 86.2 & 15.9 & 33.3 & 29.6 & 48.9  \\ 
	& MTL   & 95.5 & 85.8 & 96.6 & 74.8 & 85.8 & 68.2 & 84.5  \\   \midrule
    \multirow{1}{*}{SOTA}
        & Tripnet & {\bf 97.8} & 91.2 & {\bf 98.1} & 90.7 & 92.1 & \bf 82.7 & \bf 92.1  \\ \midrule
    \multirow{2}{*}{Ours}
    & Attention Tripnet & {\bf 97.8} & {\bf 91.5} & {\bf 98.1} & 89.7 & {\bf 92.8} & 82.1 & 92.0  \\ 
    & Rendezvous  &  97.7 & 89.4 & {\bf 98.1} & {\bf 92.0} & 92.2 & \bf 82.7 & 92.0  \\ 
    \bottomrule
    \end{tabular}
    }
\end{table*}

%% file: tables/results-perclass-v.tex
\begin{table*}[!t]
\centering
    \setlength{\tabcolsep}{7pt}
    \caption{Result Breakdown for Verb ($AP_V$) Per-Class Recognition: {\normalfont The proposed models predict correctly the most dominantly used verbs such as {\it retract, dissect, coagulate, clip}, and {\it cut} over $70.0\%$ of the time}}
    \label{tab:results:qunatitative:verb-supp}
    \resizebox{\textwidth}{!}{%
    \begin{tabular}{@{}llccccccccccr@{}
    }
    \toprule
    \multicolumn{2}{c}{Method} & Grasp & Retract & Dissect & Coagulate & Clip & Cut & Aspirate & Irrigate & Pack & Null & mAP \\ \midrule
    \multirow{3}{*}{\makecell[l]{Naive \\Baseline} }
    	& CNN  &  48.6 & 82.1 & 80.5 & 30.5 & 49.5 & 23.8 & 32.4 & 16.0 & 09.2 & 15.9 & 39.2  \\   
    	& TCN  &  24.9 & 80.2 & 66.4 & 27.4 & 31.9 & 14.7 & 14.8 & 13.9 & 2.0 & 15.4 & 29.4  \\  
    	& MTL  &  47.9 & 85.0 & 84.8 & 55.0 & 79.1 & 44.1 & 35.4 & 13.4 & 18.0 & 17.0 & 48.4  \\ \midrule
    \multirow{1}{*}{SOTA}
        & Tripnet  &  45.8 & 88.1 & 86.7 & 66.3 & 85.1 & 68.3 & 44.9 & 12.2 & 22.5 & 20.1 & 54.5  \\ \midrule
    \multirow{2}{*}{Ours}
       & Attention Tripnet & {\bf 62.4} & 89.4 & 89.4 & {\bf 69.7} & {\bf 88.5} & 84.3 & 48.5 & {\bf 20.8} & 21.4 & {\bf 22.7} & 60.2  \\ 
       & Rendezvous &  60.4 & {\bf 90.5} & {\bf 89.5} & 68.7 & 86.7 & {\bf 87.8} & {\bf 50.4} & 17.4 & {\bf 30.5} & 21.0 & {\bf 60.7}  \\ 
    \bottomrule
    \end{tabular}
    }
\end{table*}

%% file: tables/results-perclass-t.tex
\begin{table*}[!t]
\centering
    \setlength{\tabcolsep}{5pt}
    \caption{Results Breakdown for Target ($AP_T$) Per-Class Detection. \normalfont{The target ids 1..14 correspond to \textit{gallbladder, cystic-plate, cystic-duct, cystic-artery, cystic-pedicle, blood-vessel, fluid, abdominal-wall-cavity, liver, omentum, peritoneum, gut, specimen-bag, null} respectively}}
    \label{tab:results:qunatitative:target-supp}
    \resizebox{\textwidth}{!}{%
    \begin{tabular}{@{}llccccccccccccccr@{}}
    \toprule
        \multicolumn{2}{c}{\multirow{2}{*}{Method}} & \multicolumn{14}{c}{Target} & \multirow{2}{*}{mAP} \\ \cmidrule{3-16}
        \multicolumn{2}{l}{} & 1 & 2 & 3 & 4 & 5 & 6 & 7 & 8 & 9 & 10 & 11 & 12 & 13 & 14 \\ \midrule
    \multirow{3}{*}{\makecell[l]{Naive \\Baseline} }
    	& CNN  &  84.2 & 14.8 & 26.3 & 18.7 & 14.3 & 03.6 & 32.4 & 10.1 & 49.8 & 35.2 &  {\bf 08.4} & 08.4 & 69.3 & 15.9 & 28.3  \\ 
    	& TCN     &  79.9 & 10.0 & 21.4 & 19.6 & 07.0 & 01.3 & 14.8 & 06.9 & 43.1 & 27.9 & 01.9 & 09.0 & 37.4 & 15.4 & 21.4  \\ 
    	& MTL  &  85.1 & 12.2 & 29.3 & 18.6 & 06.5 & 06.4 & 30.6 & 09.8 & 55.7 &  35.8 & 02.1 & 08.4 & 71.1 & 17.5 & 28.2  \\ \midrule
    \multirow{1}{*}{SOTA}
        & Tripnet & 87.0 & {\bf 22.5} & 29.7 & 21.9 & 04.7 & 15.0 & 42.9 & 32.3 & 57.5 & 36.7 & 02.0 & {\bf 11.9} & 74.1 & 20.9 & 33.2 \\ \midrule
    \multirow{1}{*}{Ours}
    	& Attention Tripnet & 87.8 & 15.6 & {\bf 37.1} & 30.1 & 16.6 & {\bf 26.5} & 53.2 & 37.5 & {\bf 59.8} & 48.7 & 03.5 & 08.3 & {\bf 85.6} & 23.5 & 38.5  \\ 
    	& Rendezvous & {\bf 89.1} & 15.3 & 35.2 & {\bf 34.5} & {\bf 22.7} & 11.4 & {\bf 53.7} & {\bf 40.6} & 59.3 & 46.6 & 04.3 & {\bf 12.5} & 84.0 & {\bf 25.0} & 38.3  \\
    	\bottomrule
    \end{tabular}
    }
\end{table*}

%% file: tables/results-perclass-iv.tex
\begin{table*}[!t]
\centering
    \setlength{\tabcolsep}{6pt}
    \caption{Top-10 Predicted Instrument-Verb  ($AP_{IV}$) Association: {\normalfont  The proposed models show a higher capability of detecting the top combinations that represent the most likely usage pattern of the individual instrument class, as well as the most clinical relevant instrument roles}.}
    \label{tab:results:qunatitative:topIV-supp}
    \resizebox{\columnwidth}{!}{%
    \begin{tabular}{@{}lrrcclrrcclrr@{}}\toprule
        \multicolumn{3}{c}{Tripnet} & \phantom{abc} &\phantom{abc} & \multicolumn{3}{c}{Attention Tripnet} & \phantom{abc} &\phantom{abc} & \multicolumn{3}{c}{Rendezvous} \\ 
        \cmidrule{1-3} \cmidrule{6-8} \cmidrule{10-13}
        {Triplet} &\phantom{abc}& {AP} &&& {Triplet} &\phantom{abc}& {AP} &&& {Triplet} &\phantom{abc}& {AP} \\ \midrule 
        bipolar,coagulate && 88.71 & & & grasper,retract && 90.29 & & & grasper,retract && 90.51 \\
        grasper,retract && 87.58 &&  & hook,dissect && 87.18 & & & hook,dissect && 90.38 \\
        hook,dissect && 86.88 & & & bipolar,coagulate && 78.17 & & & bipolar,coagulate && 88.05 \\
        scissors,cut && 68.93 & & & scissors,cut && 77.99 & & & scissors,cut && 86.40 \\
        clipper,clip && 67.10 & & & clipper,clip && 70.66 & & & clipper,clip && 82.65 \\
        irrigator,aspirate && 57.51 & & & irrigator,aspirate && 57.10 & & & irrigator,aspirate && 51.95 \\
        grasper,grasp && 23.54 & & & grasper,grasp && 37.1 & & & grasper,grasp && 48.97 \\
        irrigator,null-verb && 16.28 & & & irrigator,dissect && 24.49 & & & grasper,null-verb && 28.91 \\
        clipper,null-verb && 16.10 & & & grasper,null-verb && 20.81 & & & scissors,null-verb && 21.68 \\
        grasper,null-verb && 12.47 & & & irrigator,irrigate && 16.47 & & & irrigator,dissect && 20.42 \\
        \midrule
        mean && 52.51 & & & && 56.03 &  && && 60.99 \\
    \bottomrule
    \end{tabular}
    }
\end{table*}

%% file: tables/results-perclass-it.tex
\begin{table*}[!t]
\centering
    \setlength{\tabcolsep}{6pt}
    \caption{Top-10 Predicted Instrument-Target ($AP_{IT}$) : {\normalfont  The proposed models predict well the clinically most relevant situations, which are {\it clipping} and {\it cutting} of {\it cystic-artery} and {\it cystic-duct} among their top detected instrument-target labels.}}
    \label{tab:results:qunatitative:topIT-supp}
    \resizebox{\columnwidth}{!}{%
    \begin{tabular}{@{}lrcclrcclr@{}}\toprule
        \multicolumn{2}{c}{Tripnet} & \phantom{abc} & \phantom{abc} & \multicolumn{2}{c}{Attention Tripnet} & \phantom{abc} & \phantom{abc} & \multicolumn{2}{c}{Rendezvous} \\ 
        \cmidrule{1-2} \cmidrule{5-6} \cmidrule{9-10}
    {Triplet} & {AP} &&& {Triplet} & {AP} &&& {Triplet} & {AP} \\ \midrule 
    grasper,gallbladder & 82.49 & & & grasper,gallbladder & 83.65 & & & grasper,gallbladder & 89.96 \\
    grasper,specimen-bag & 76.50 & & & grasper,specimen-bag & 82.34 & & & grasper,specimen-bag & 81.75 \\
    bipolar,liver & 67.09 & & & bipolar,liver & 68.37 &&  & hook,gallbladder & 76.20 \\
    hook,gallbladder & 57.96 & & & hook,gallbladder & 63.82 & & & grasper,liver & 66.22 \\
    irrigator,fluid & 57.51 & & & grasper,liver & 54.26 & & & bipolar,liver & 62.45 \\
    clipper,cystic-artery & 47.44 & & & irrigator,cystic-pedicle & 49.38 & & & clipper,cystic-duct & 59.68 \\
    grasper,liver & 44.52 & & & scissors,cystic-artery & 48.44 & & & bipolar,blood-vessel & 57.18 \\
    scissors,cystic-duct & 42.57 & & & grasper,omentum & 47.20 & & & grasper,omentum & 54.98 \\
    scissors,cystic-artery & 40.37 & & & irrigator,fluid & 47.11 & & & scissors,cystic-artery & 53.84 \\
    clipper,cystic-duct & 39.62 & & & bipolar,abdominal-wall-cavity & 46.07 & & & irrigator,fluid & 51.95 \\
    \midrule 
    mean & 55.61 & & & & 59.06 & & &  & 65.42 \\
    \bottomrule
    \end{tabular}
    }
\end{table*}

%% file: tables/results-ablation-sequence.tex
\begin{SCtable}[][ht]
\setlength{\tabcolsep}{6pt}
\caption{Ablation Study on Attention Sequence Modeling: {\normalfont This study shows the usefulness of our class-wise mapping over the contemporally patch-base sequence in Vision Transformer \citep{dosovitskiy2020image}}}
\label{tab:results:qunatitative:ablation:sequence-supp}
\resizebox{0.5\columnwidth}{!}{%
\begin{tabular}
    {@{}lrcccc@{}}\toprule
    Model && Layer Size & AP$_{IV}$ & AP$_{IT}$ & AP$_{IVT}$ \\ \midrule
    Patch-base sequence && 6 & 33.4 & 29.3 & 24.1 \\
    Class-wise mapping  && 2 & 36.0 & 34.1 & 27.0 \\
    Class-wise mapping  && 8 & \bf 39.4 & \bf 36.9 & \bf 29.9 \\
    \bottomrule
\end{tabular}
}
\end{SCtable}

%% file: tables/results-ablation-auxilliary.tex
\begin{SCtable}[][ht]
    \setlength{\tabcolsep}{15pt}
    \caption{Ablation Study on Use of Auxiliary Loss: {\normalfont This study shows that learning the individual components of the triplets in the same network pipeline helps the model to better understand the triplets.}}
    \label{tab:results:qunatitative:ablation:aux-loss-supp}
    
    \resizebox{0.5\columnwidth}{!}{%
    \begin{tabular}
        {@{}lrccc@{}}\toprule
        Model &&  AP$_{IV}$ & AP$_{IT}$ & AP$_{IVT}$ \\ \midrule
        Without aux-loss &&  33.6 & 27.0 & 21.2  \\
        With aux-loss  &&  \bf 36.0 & \bf 34.1 & \bf 29.9 \\
        \bottomrule
    \end{tabular}
    }
\end{SCtable}